\documentclass[conference]{IEEEtran}
\IEEEoverridecommandlockouts

\usepackage{cite}
\usepackage{amsmath,amssymb,amsfonts}
\usepackage{algorithmic}
\usepackage{graphicx}
\usepackage{textcomp}
\usepackage{xcolor}
\usepackage{subcaption}
\usepackage{wrapfig}

\DeclareMathOperator*{\argmax}{arg\,max}

\def\BibTeX{{\rm B\kern-.05em{\sc i\kern-.025em b}\kern-.08em
    T\kern-.1667em\lower.7ex\hbox{E}\kern-.125emX}}
\begin{document}

\title{Assessing Deep Neural Networks as \\Probability Estimators}

\author{\IEEEauthorblockN{Yu Pan\textsuperscript{1}, Kwo-Sen Kuo\textsuperscript{2}, Michael L. Rilee\textsuperscript{2}, Hongfeng Yu\textsuperscript{1}}
\IEEEauthorblockA{
$^1$University of Nebraska-Lincoln, Lincoln, NE, USA\\
$^2$Bayesics, LLC, Bowie, MD, USA
}
}

\maketitle

\begin{abstract}
Deep Neural Networks (DNNs) have performed admirably in classification tasks. However, the characterization of their classification uncertainties, required for certain applications, has been lacking. In this work, we investigate the issue by assessing DNNs’ ability to estimate conditional probabilities and propose a framework for systematic uncertainty characterization. Denoting the input sample as $x$ and the category as $y$, the classification task of assigning a category $y$ to a given input $x$ can be reduced to the task of estimating the conditional probabilities $p(y|x)$, as approximated by the DNN at its last layer using the softmax function. Since softmax yields a vector whose elements all fall in the interval (0, 1) and sum to 1, it suggests a probabilistic interpretation to the DNN's outcome. Using synthetic and real-world  datasets, we look into the impact of various factors, e.g., probability density $f(x)$ and inter-categorical sparsity, on the precision of DNNs' estimations of $p(y|x)$, and find that the likelihood probability density and the inter-categorical sparsity have greater impacts than the prior probability to DNNs' classification uncertainty.
\end{abstract}


\section{Introduction}

The potential of deep neural networks (DNNs) has been amply demonstrated with classification. If we denote the input sample as a vector $x$ and an $L$-category vector as $y = (y_1, ...,y_L)$, a classification task can be viewed as taking the following two steps: 1) it first predicts the conditional probability $p(y|x)$, and then 2) makes a decision on the category an input sample belongs to based on some specific criteria, such as $y=\argmax_yp(y|x)$, i.e. identifying with the largest element of $y$. Although the final classification result is of primary interest, the intermediate result $p(y|x)$ is necessary for scientific applications, in which the characterization of classification uncertainties is desired. However, there is a lack of systematic investigations into this characterization.

A DNN often uses softmax on the output of its last layer. Since softmax yields a vector whose elements all fall in the interval (0,1) and sum to 1, it suggests a probabilistic interpretation of the DNN's outcome and is used to approximate the probability of the categorical variable $y$. Typically during training, the labels of all input samples fed into the DNN are in the one-hot format (i.e., each sample is associated with a single category). The DNN learns $p(y|x)$ \emph{implicitly} by minimizing the cross-entropy between the output and the one-hot label without revealing $p(y|x)$. The main mechanism for the DNN to capture $p(y|x)$ is through relating local samples of $x$ and the frequencies of $y$. The (local) sparsity of $x$ in the training dataset, therefor, may limit the capability of the DNN to capture $p(y|x)$.

We are interested in assessing the  quality of the prediction of $p(y|x)$ and exploring potential factors that may impact the performance metric. However, the lack of ground truth makes it difficult to assess the prediction of $p(y|x)$ generated by DNNs. In this paper, we address these challenges with the following main contributions:
\begin{itemize}
  \item We propose an innovative generative framework with two paths: one for directly inferring $p(y|x)$ assuming Gaussian probability density functions (pdf's) and one for generating and training a DNN to approximate $p(y|x)$.
  \item  We conduct extensive and systematic experiments for both 1-D and high-dimensional inputs to gain insights that suggest the likelihood probability density and the inter-categorical sparsity are the more influential factors on the performance metric than the prior probability.
\end{itemize}



\section{Related Work}

We describe works related to ours in this section. We note that the sample labeling process naturally biases the distribution because, under the one-hot convention, existing works tend to ignore the samples that are unsure of. We cannot uncritically assume that the distributions of the labeled samples, regardless of training or testing, accurately represent those of the population in the real world.

\subsection{Estimating Probability using DNNs}

Substantial work has been conducted to estimate the underlying probability in training data using DNNs. Based on how the actual probability (density) function is approximated, we may categorize the existing work into two categories: implicit and explicit estimations.


When a model uses implicit estimation, it does not approximate the distribution in a closed form but usually generates samples subject to the distribution. Generative Adversary Network (GAN)~\cite{goodfellow2014generative} consists of a generator and a discriminator, which co-evolve to achieve the best performance. The generator implicitly models the distribution of training data, and the discriminator attempts to differentiate between the true distribution and the synthesized distribution from the generator. The generator, however, has not been leveraged to generate samples with prescribed distributions for uncertainty characterization. Ever since its invention, GAN has evolved into a large family of architectures
\cite{mirza2014conditional, arjovsky2017wasserstein, brock2018large, isola2017image, karras2019style, radford2015unsupervised, zhang2017stackgan, zhang2019self, zhu2017unpaired}.


Explicit estimation attempts to learn the distribution $p(x|y)$ in a closed form. Some pioneering studies~\cite{lu2020universal, achille2018emergence, magdon1999neural} discuss the capability of DNNs to approximate probability distributions. Mixture Density Network (MDN)~\cite{bishop1994mixture} predicts not only the expected value of a target but also the underlying probability distribution. Given an input, MDN extends maximum likelihood by substituting Gaussian pdf with the mixture model. Probabilistic Neural Network (PNN)~\cite{specht1990probabilistic} uses a Parzen window to estimate the probability density for each category $p(x|y)$ and then uses Bayes' rule to calculate the posterior $p(y|x)$. PNN is non-parametric in the sense that it does not need any learning process, and at each inference, it uses all training samples as its weights. These techniques do not seem to consider the possibility that the distributions of the labeled samples may deviate from those of the population.

\subsection{Approximate Inference}

As the inference process for complex models is usually computationally intractable, one cannot perform inference exactly and must resort to approximation. Approximate inference methods may also be categorized into two categories: sampling and variational inference.


Sampling is a common method to address intractable models. One can draw samples from a model and fit an approximate probabilistic model from the samples. There are classic sampling methods, such as inverse sampling, rejection sampling, and Gibbs sampling, as well as more advanced sampling methods, such as Markov chain Monte Carlo (MCMC) \cite{geyer1992practical}. Our framework is similar to the sampling method in the sense that it generates samples for training and testing a DNN model.


Variational inference is an alternative to sampling methods. It approximates the original distribution $p(x|y)$ with a fit $q(x|y)$,  turning it  into an optimization problem. Accordingly, variational autoencoder~\cite{kingma2013auto}  approximates the conditional distribution $p(z|x)$ of latent variables $z$ given an input $x$ using a function $q(z|x)$ by reducing the KL-Divergence between the two distributions. This results in an additional loss component and a specific estimator for the training algorithm called the Stochastic Gradient Variational Bayes (SGVB) estimator. Researchers have incorporated some more sophisticated posteriors $q(z|x)$ to extend variational autoencoder~\cite{rezende2014stochastic, nalisnick2016approximate, kingma2016improving}.

\subsection{Bayesian Neural Networks}

Since deep learning methods typically operate as black boxes, the uncertainty associated with their predictions is challenging to quantify. Bayesian statistics offer a formalism to understand and quantify the uncertainty associated with DNN predictions. Differing from point estimation used in a typical training process such as Stochastic Gradient Descent, Bayesian neural networks learn a probability distribution over the weights~\cite{jospin2020hands, hernandez2015probabilistic, mackay1995bayesian, springenberg2016bayesian}.

\subsection{Calibration method}

Guo et al. \cite{guo2017calibration} translates the problem of calibrating the prediction made by a neural network into counting how many correct predicted samples are made. Their calibration depends on specific dataset whereas we adopt a different calibration mechanism in which our generative model can generate training datasets according to different hyper parameters.

\section{Framework}

\begin{figure}[t]
    \centering
    \includegraphics[width=0.9\linewidth]{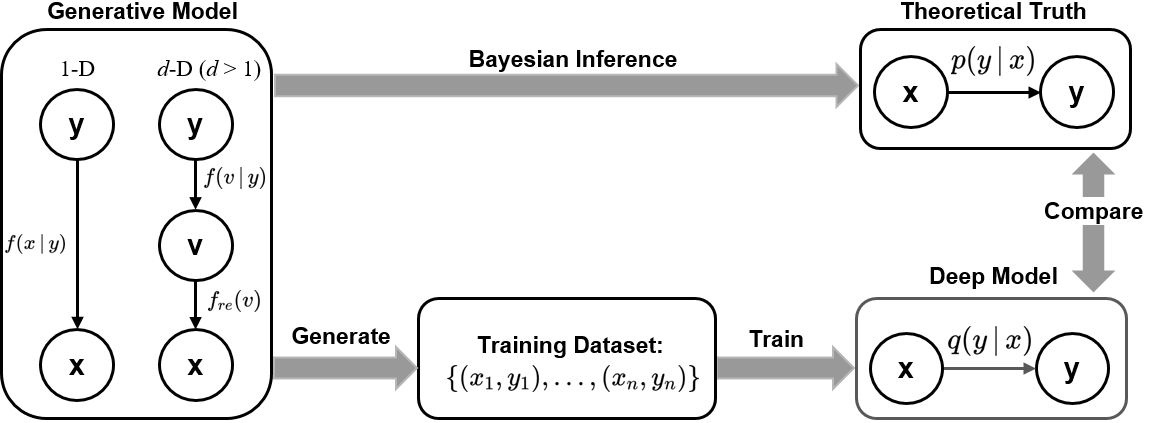}
    \caption{\small{Two paths of our assessment framework. A dataset generator is constructed and used to generate synthetic samples. Along the fristpath of Bayesian inference, it is easy to infer the posterior $p(y|x)$ from prescribed prior $p(y)$ and likelihood $p(x|y)$. Along the path of sampling and training, it first samples the class label $y=y_i$ based on a prescribed discrete distribution of $y$. For 1-D, $x_i$ is sampled according to a prescribed likelihood  $f(x|y=y_i)=\mathcal{N}(\mu(y_i),\sigma(y_i))$. For $d>1$ dimension, we first sample a vector $\mathbf{v}$ according to a prescribed Gaussian pdf $f(\mathbf{v}|y=y_i)=\mathcal{N}(\boldsymbol{\mu}(y_i),\boldsymbol{\Sigma}(y_i))$ in the reduced dimension, which is 2-D for the case of this study. We then map this 2-D vector $\mathbf{v}$ to a $d$-dimensional vector $\mathbf{x}$ using our reconstructive mapping and add the $(x_i,y_i)$ pair to the training dataset. Repeating this $n$ times, we generate a training dataset $\mathcal{D}$ containing $n$ data samples, which we feed into the DNN for training. When the DNN is fully trained, its predicted probability $q(y|x)$ given any input $x$ can be compared with the ground truth $p(y|x)$.}}
    \label{fig:framework}
\end{figure}

To assess how well a DNN captures the posterior pdf $p(y|x)$ embedded in a dataset, we must first know the ``truth" of the dataset. Yet, given an arbitrary dataset for a typical classification task, it is challenging to estimate the ground truth of the conditional relationship. We end up with a chicken-and-egg situation: We need the ``ground truth" to evaluate an estimate, but we can only approximate the ground truth with an estimate. It thus becomes impossible to characterize the classification uncertainty with confidence.


To address this problem, we introduce a new assessment framework to systematically characterize DNNs' classification uncertainty, as illustrated in Fig.~\ref{fig:framework}. The key idea of our framework is to construct a data generative model by specifying all the information required for the estimation, including the prior distribution $p(y)$ and the dependency of $x$ on $y$, thus establishing the ground truth. We then proceed along two paths: 1) The first path is through Bayesian inference, in which we directly calculate $p(y|x)$ through Bayes theorem, i.e. $p(y|x)\propto p(x|y)p(y)$; and 2) the second path is through generating a dataset using the aforementioned generative model and then training a DNN-based classifier $q(y|x)$ as an approximation to the ``true" $p(y|x)$ and evaluated for its probability approximation ability. The second path is similar to approximate inference by sampling. After the DNN is fully trained, we can compare how close the results of these two paths are, or to say, we can compare the prediction made by the DNN and the ``ground truth" from our dataset generator.

In practice, it is non-trivial to directly estimate high-dimensional distributions for many real-world cases. To tackle these cases, we first apply a dimensionality reduction technique, if necessary, to map the high-dimensional input samples to a more manageable low-dimensional space, from which we construct a generative model to generate an extensive set of synthetic samples by densely sampling the reduced-dimensionality space and inversely mapping back to the original high-dimensional space (aka reconstructive mapping). We can thus now sample this extensive dataset of synthetic samples according to prescribed prior $p(y)$ and likelihood $p(x|y)$ to serve as ground truth.

%

\subsection{Framework Formalization}
\label{sec:framework_formalization}

A generative model produces an extensive dataset of synthetical samples, which we sample according to some prescribed prior $p(y)$ and likelihood $p(x|y)$ to serve as ``ground truth". For 1-D, the likelihood $p(x|y)$ is represented by a Gaussian pdf $f(x|y)=\mathcal{N}(\mu(y),\sigma(y))$. For $d>1$ dimension, we assume $x$ is embedded in a lower dimensional manifold for many real world datasets. Thus, the likelihood can be represented by a composite of lower-dimensional Gaussian pdf's $f(\mathbf{v}|y)=\mathcal{N}(\boldsymbol{\mu}(y),\boldsymbol{\Sigma}(y))$ and through a reconstructive mapping function $f_{re}$.

For the real-world MNIST dataset studied (see Section \ref{sec:hd_case}), we find that x stays essentially on a 2-D manifold, so we have $\mathbf{v} \in \mathbb{R}^2$ and a reconstructive mapping function $f_{re}:\mathbf{V} \subset \mathbb{R}^2 \rightarrow \mathbf{X} \subset \mathbb{R}^d$. Here any bijective function that maps a 2-D vector $\mathbf{v}$ back to a $d$-dimensional vector $\mathbf{x}$ will work as a reconstructive mapping and $f_{re}$ can be seen as a decoding mapping. We detail our investigations into both the 1-D and high-dimensional cases in Section \ref{sec:experiment}.

\subsection{Two Paths}


We detail the two paths of inference used in our framework in the following subsections.

\subsubsection{Bayesisan Inference}
\label{sec:bayesian_inference}

Since the synthetic samples produced by the generative model are constrained by prescribed prior $p(y)$ and likelihood $p(x|y)$, we can easily infer $p(y|x)$ based on Bayesian rule for the 1-D case:
\begin{equation}
\small{
p(y|x)= \frac{f(x|y)p(y)}{f(x)}=\frac{f(x|y)p(y)}{\sum_y{f(x|y)p(y)}}
}
\end{equation}
For $d$-dimension ($d>1$), we may use $f(\mathbf{v}|y)$ and $p(y)$ to infer $p(y|\mathbf{x})$:
\begin{equation}
\small{
p(y|\mathbf{x}) = p(y|\mathbf{v}) = \frac{f(\mathbf{v}|y)p(y)}{f(\mathbf{v})}
}
\end{equation}
Appendix~\ref{sec:proof} gives the detailed mathematical proof.

\subsubsection{Data Sampling and Training}

We sample, in both 1D and d>1, class labels based on a prescribed prior of $p(y)$.
A 1-D training dataset is generated according to a prescribed likelihood $f(x|y=y_i)$, where $y_i$ is a sample of $y$, with the likelihood assumed to be Gaussian, i.e., $f(x|y=y_i)=\mathcal{N}(\mu(y_i),\sigma(y_i))$.
For $d$-dimension, we first sample, according to a 2-D Gaussian pdf distribution $f(\mathbf{v}|y=y_i)=\mathcal{N}(\boldsymbol{\mu}(y_i),\boldsymbol{\Sigma}(y_i))$, a 2-D vector $\mathbf{v}_{i}$ which we map to a $d$-dimensional vector $\mathbf{x}_i$.
Subsequently, we add $(x_i,y_i)$ to the training dataset. Repeating this $n$ times, we generate a training dataset $\mathcal{D}$ containing $n$ data samples, which are then fed into a DNN for training. When the DNN is fully trained, we compare the predicted probability $q(y|x)$ of any given input $x$ with the ground truth $p(y|x)$.

\subsection{Comparison}
\label{sec:comparison}

For each configuration of the generative model, we have an inferred $p(y|x)$ and a trained (predicted) $q(y|x)$. We can compare these two conditional probabilities at each sampled point of $x$. More importantly, we can systematically explore all possible configurations of the generative model and find the main factors affecting the approximation precision of $q(y|x)$. Given the complexity of this exploration, here we report more comprehensively on only 1-D cases. For $d$-dimensional cases, we use the MNIST dataset, fit its 2-D representation as a Gaussian mixture, and explore this specific configuration.

\section{Experiment and Evaluation}
\label{sec:experiment}

\begin{figure}[t!]
\begin{center}
$\begin{array}{c@{\hspace{0.05\linewidth}}c}
\includegraphics[width = 0.45\linewidth]{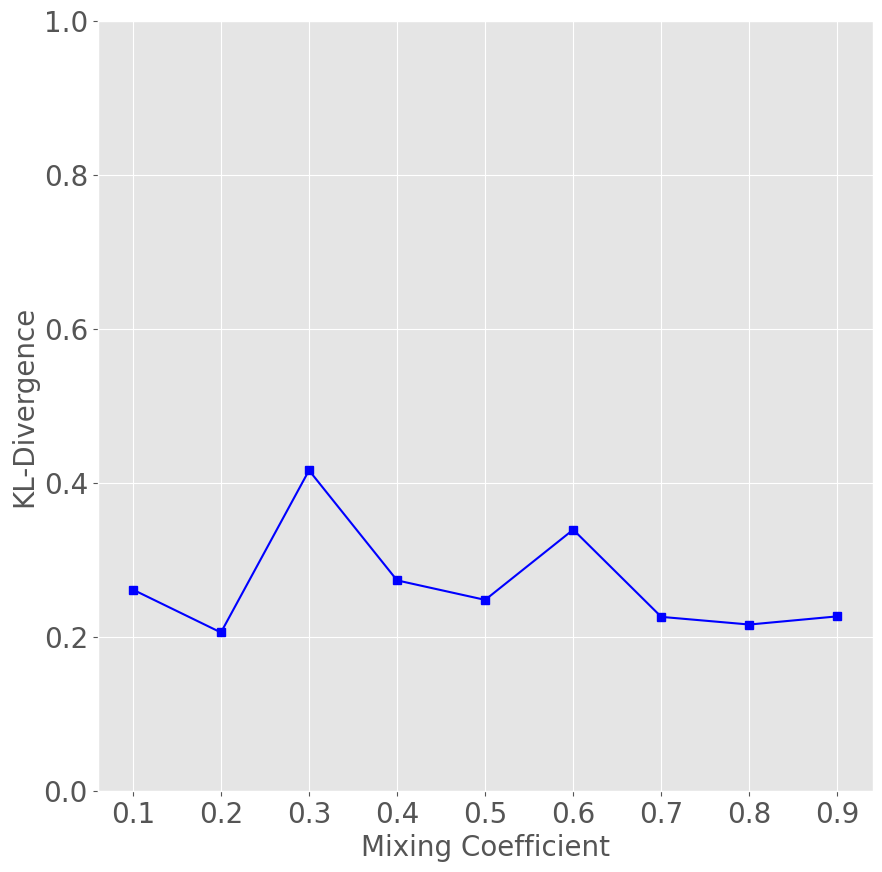} &
\includegraphics[width = 0.45\linewidth]{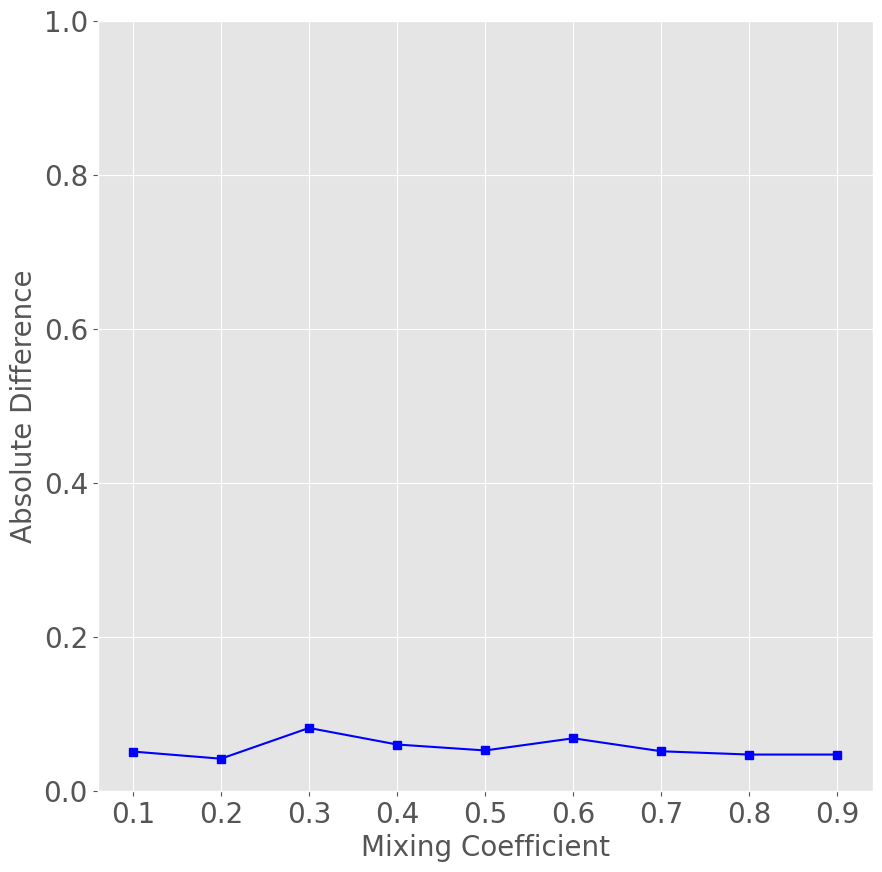}\\
\mbox{\small{(a)}} & \mbox{\small{(b)}} \\
\includegraphics[width = 0.45\linewidth]{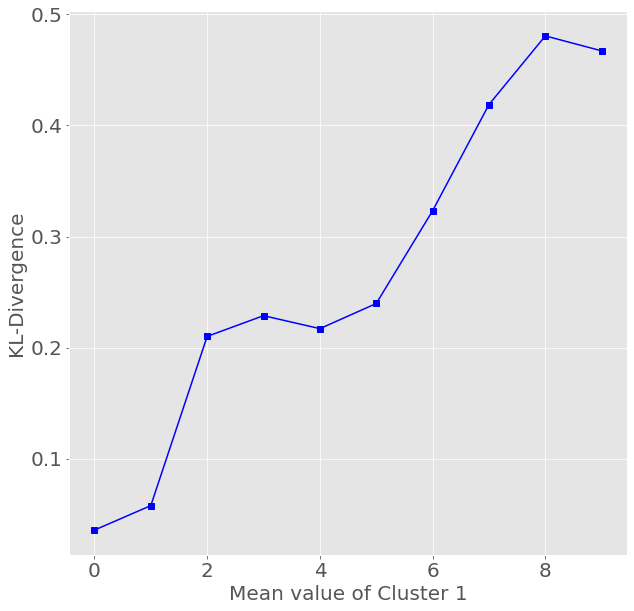} &
\includegraphics[width = 0.45\linewidth]{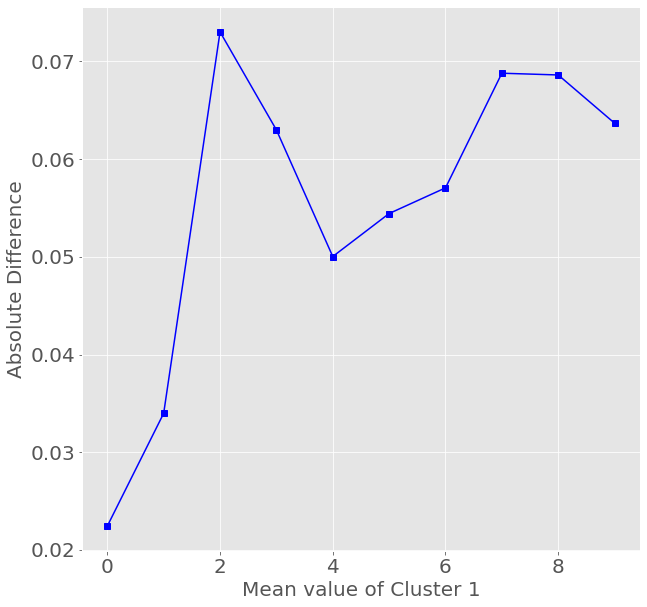}\\
\mbox{\small{(c)}} & \mbox{\small{(d)}}\\
\includegraphics[width = 0.45\linewidth]{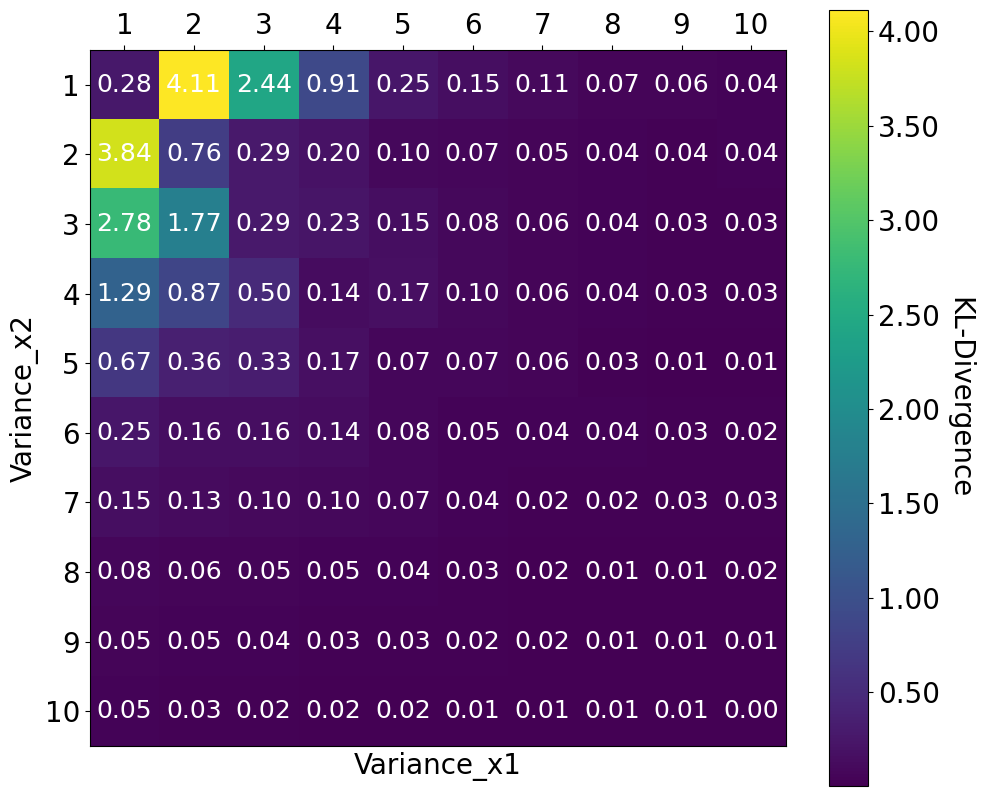} &
\includegraphics[width = 0.45\linewidth]{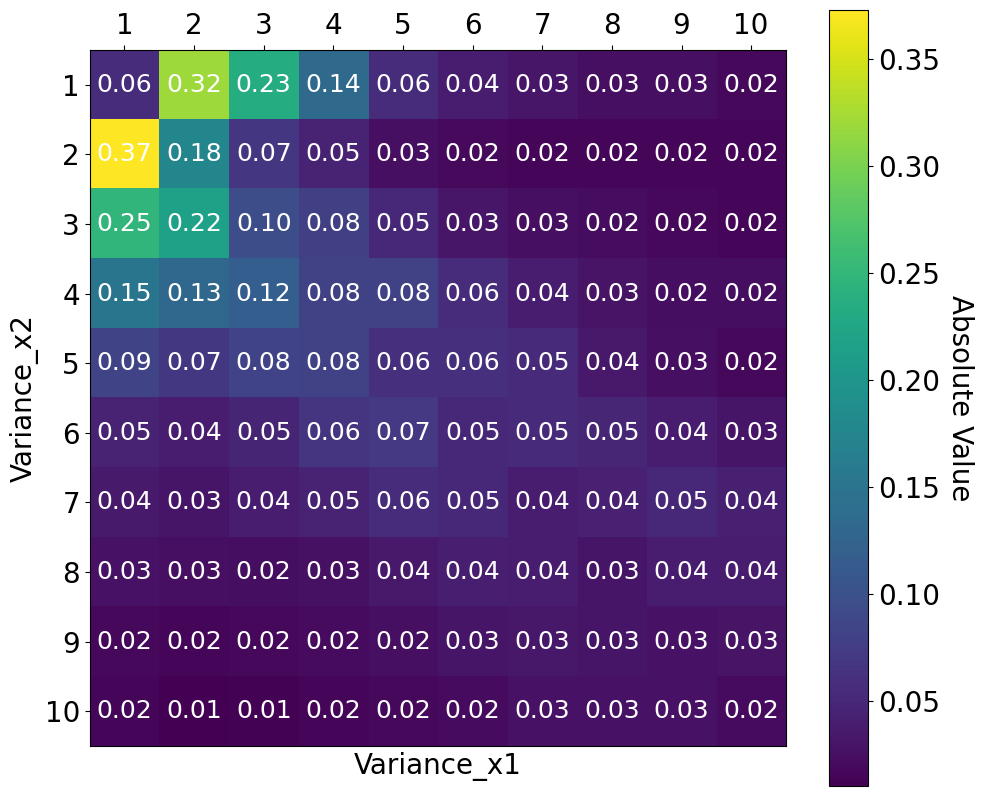}\\
\mbox{\small{(e)}} & \mbox{\small{(f)}}
\end{array}$
\end{center}
\vspace{-0.15in}
\caption{\small{Prediction precision as the function of various factors. (a) plots the prediction precision in KL-Divergence as the function of the mixing coefficient $p(y=1)$. (c) plots the prediction precision in KL-Divergence as the function of the mean value of cluster 1, $\mu_1$. (e) plots the prediction precision in KL-Divergence as the function of variances $\sigma_1$ and $\sigma_2$. (b), (d), and (f) plot the counterparts of (a), (c), and (e) in Absolute Difference, respectively.}}
\label{fig:1d_grid_search}
\end{figure}

We use Google Colab\cite{bisong2019google} with Nvidia T4 and P100 GPUs to run our experiment. We discuss below the experiment setups and results for 1-D and high dimensional cases, respectively.
\subsection{One Dimensional Case}
\label{sec:1d_experiment}

To simplify our experiment without loss of generality, we use a mixture of two Gaussian pdf's as our data generator. Here, we call each Gaussian pdf a \emph{cluster}. The 1-D generative model can be parameterized as $f(x)=\sum_{k=1}^{2} p(y=k)\phi_k(x)$, where $\sum_{k=1}^{2} p(y=k)=1$, $\phi_k(x)=\mathcal{N}(\mu_k,\sigma_k)$, and $p(y=k)$, $\mu_k$, and $\sigma_k$ are the generative model parameters.

\subsubsection{Systematic Analysis}
\label{sec:systematic_analysis}

\begin{figure}[t!]
\begin{center}
$\begin{array}{c@{\hspace{0.05\linewidth}}c}
\includegraphics[width = 0.45\linewidth]{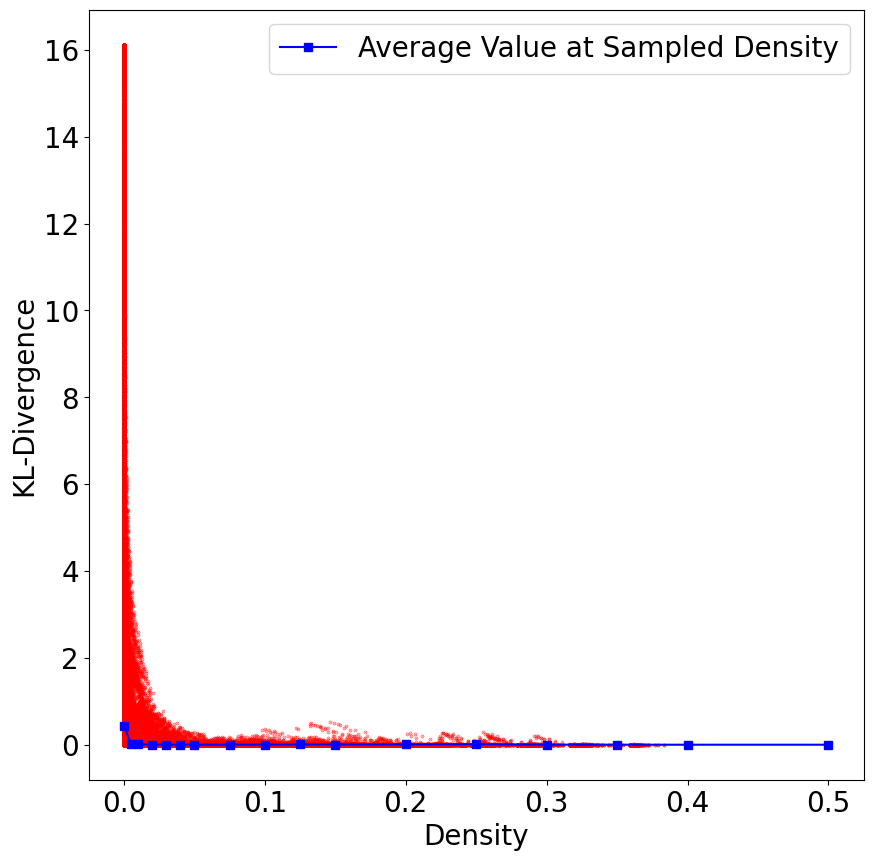} &
\includegraphics[width = 0.45\linewidth]{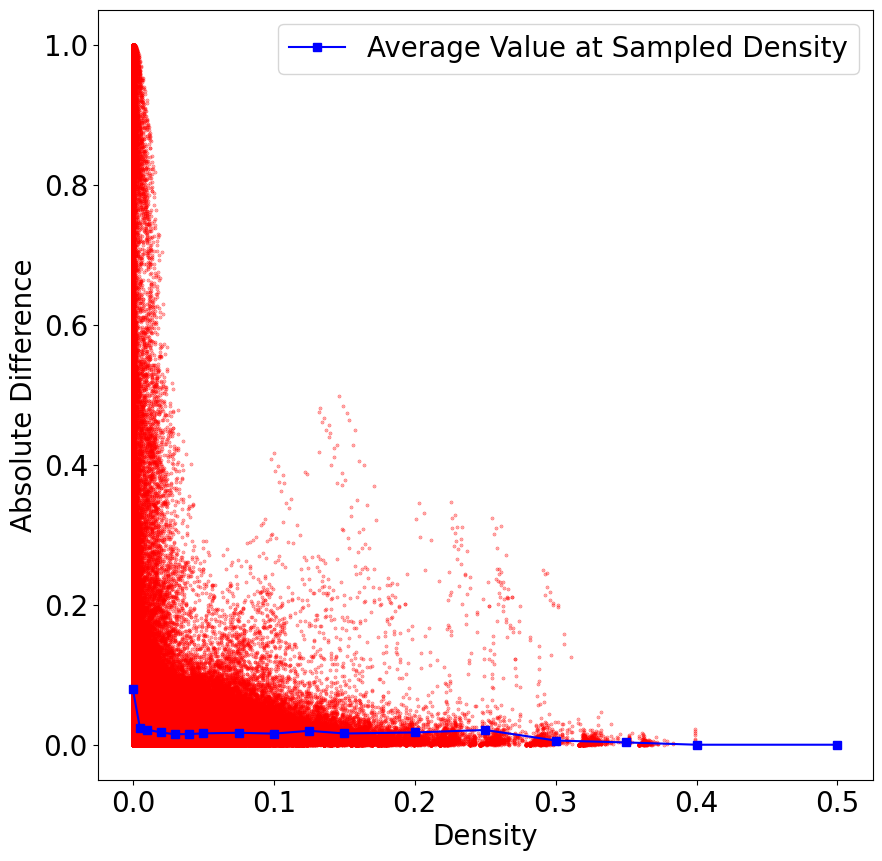}\\
\mbox{\small{(a)}} & \mbox{\small{(b)}} \\
\includegraphics[width = 0.45\linewidth]{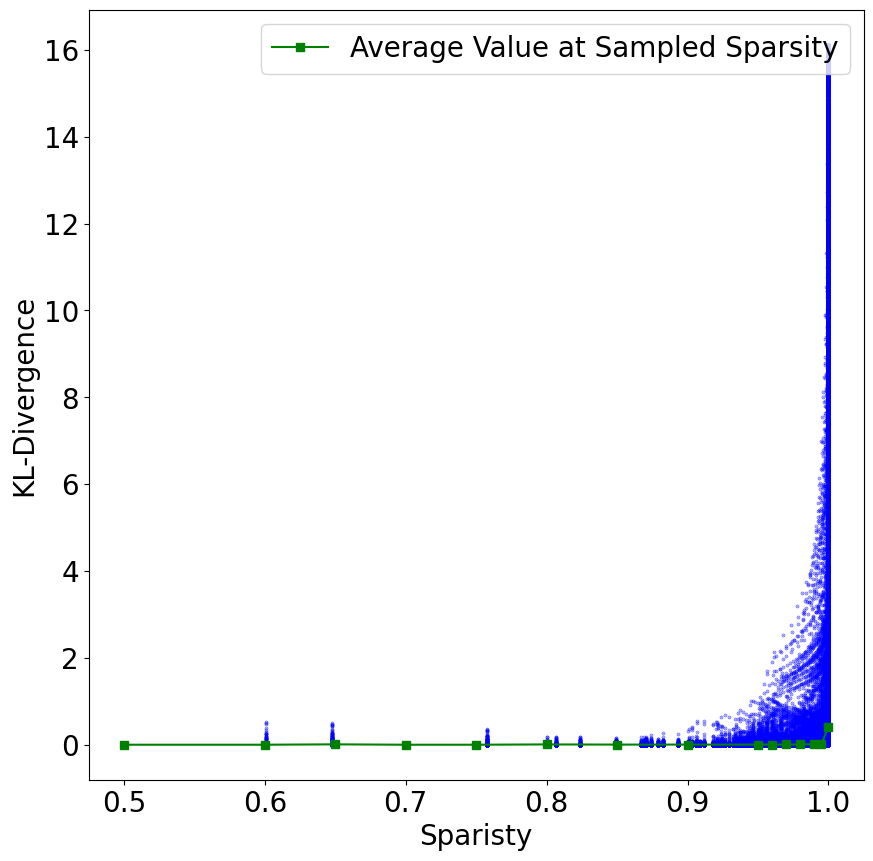} &
\includegraphics[width = 0.45\linewidth]{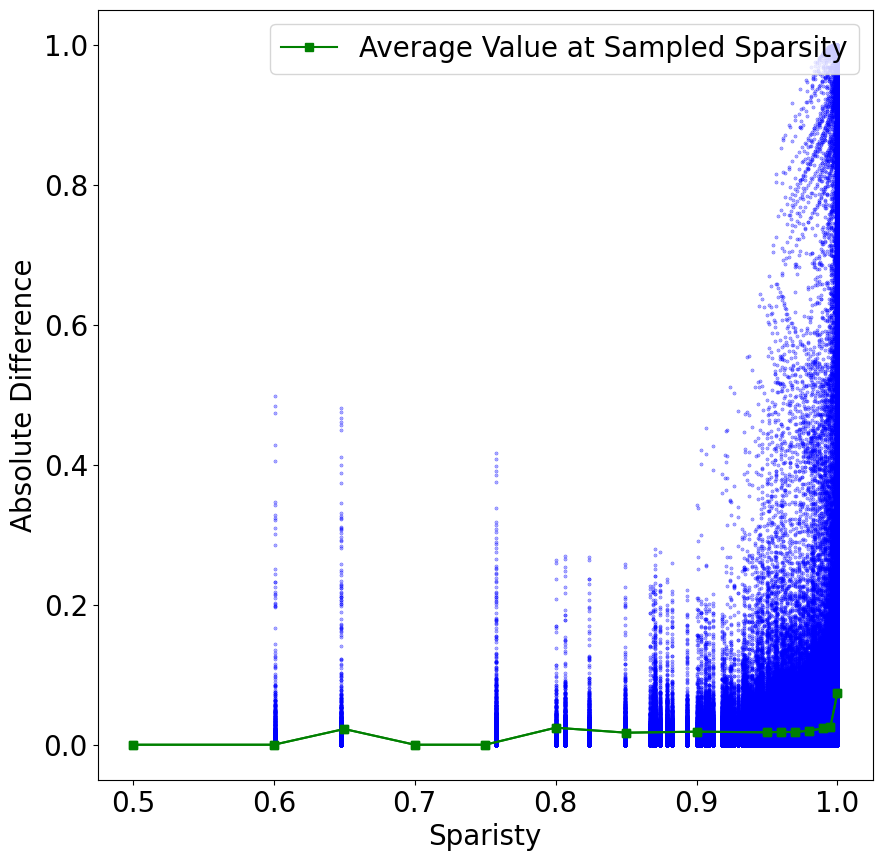}\\
\mbox{\small{(c)}} & \mbox{\small{(d)}}
\end{array}$
\end{center}
\vspace{-0.15in}
\caption{\small{Prediction precision as the function of the marginal density and sparsity. Each point represents the situation for one sampled point $x$, among all the sampled points and all the parametric configurations of our generative model. (a) plots the prediction precision in KL-Divergence as the function of density. (b) plots the prediction precision in Absolute Difference as the function of density. (c) plots the prediction precision in KL-Divergence as the function of sparsity. (d) plots the prediction precision in Absolute Difference as the function of sparsity. For each figure, we also divide the x axis into several bins and calculate the average value for each bin. Then we overlay the average values as the function of sampled bins on each figure.}}
\label{fig:1d_density_sparsity}
\end{figure}

\paragraph{Grid Search in Parametric Space}
To explore how each parameter would influence the DNN prediction, we first conduct a grid search in the parametric space $p(y=k)$, $\mu_k$, and $\sigma_k$ of the generative model, where each grid point is a combination of different values of $p(y=k)$, $\mu_k$, and $\sigma_k$. Here, we sample $p(y=1) \in [0.1,0.9]$ with step size of $0.1$, and $\mu_1 \in [0,9]$ with a step size of $1$ subject to the condition $\mu_1+\mu_2=0$. We also sample both $\sigma_1$ and $\sigma_2 \in [1,10]$ with a step size of $1$. Therefore, our parametric space is essentially a space containing $9\times 10\times10\times10=9000$ grid points, which means we generate a training set and then train a fully connected DNN $9000$ times. For each grid point (i.e., a configuration of the generative model), we generate $10000$ data samples and labels in a stochastic (random) manner. After the DNN is fully trained, we get the prediction precision of the trained DNN at the sampled $x \in [-35,35]$ with a step size $0.5$. We choose 35 as it is slightly larger than $max(\mu) + 2.5 max(\sigma)$ and can cover the majority area of non-trivial density $f(x)$, where $max(\mu)=9$ and $max(\sigma)=10$ are the maximum values of $\mu$ and $\sigma$, respectively. Then, we calculate the mean prediction precision at these $142$ sampled points. When we plot the prediction precision as a function of each of these three factors, we marginalize the other two factors.

To measure the prediction precision, we use two metrics, KL-Divergence:
\begin{equation}
\label{eq:kl}
\small{
D_{KL}(p(y|x) || q(y|x))=-\sum_{k=1}^{2} p(y=k)\log(\frac{q(y=k|x)}{p(y=k|x)})
}
\end{equation}
and Absolute Difference:
\begin{equation}
\label{eq:abs}
\small{
D_{ABS}(p(y|x),q(y|x))=|p(y=1|x)-q(y=1|x)|
}
\end{equation}

Fig. \ref{fig:1d_grid_search} shows the prediction precision as the function of mixing coefficient $p(y=1)$, the mean value of cluster 1 $\mu_1$, and the variances of two clusters $\sigma_1$ and $\sigma_2$. We can conclude from Fig. \ref{fig:1d_grid_search} that there is only a marginal relationship between the mixing coefficient and the prediction precision. Increasing the distance between the cluster mean values can improve the prediction precision, which we discuss below. The impact of variances is relatively complex. Generally, we see two trends: prediction accuracy decreases with both smaller variance values and with decreasing distance between the two variances. The only exception for the second trend is that, when both variances are equal, the prediction is more accurate than expected from the second trend.

\begin{figure}[t!]
\begin{center}
$\begin{array}{c@{\hspace{0.03\linewidth}}c}
\includegraphics[width = 0.48\linewidth]{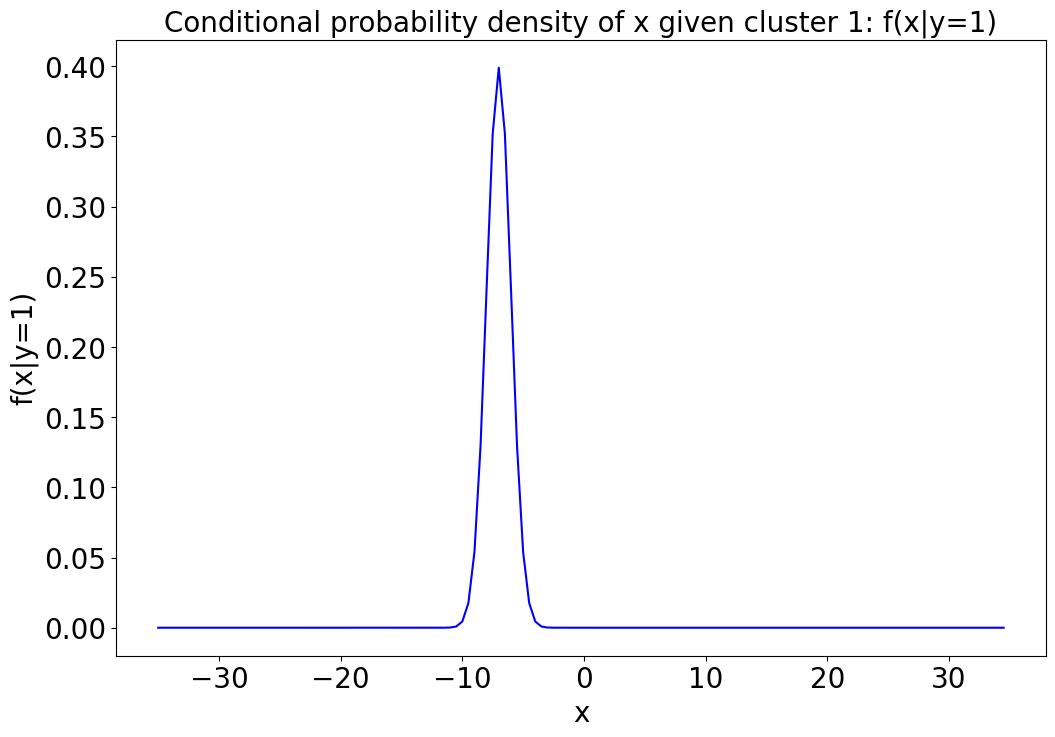} &
\includegraphics[width = 0.48\linewidth]{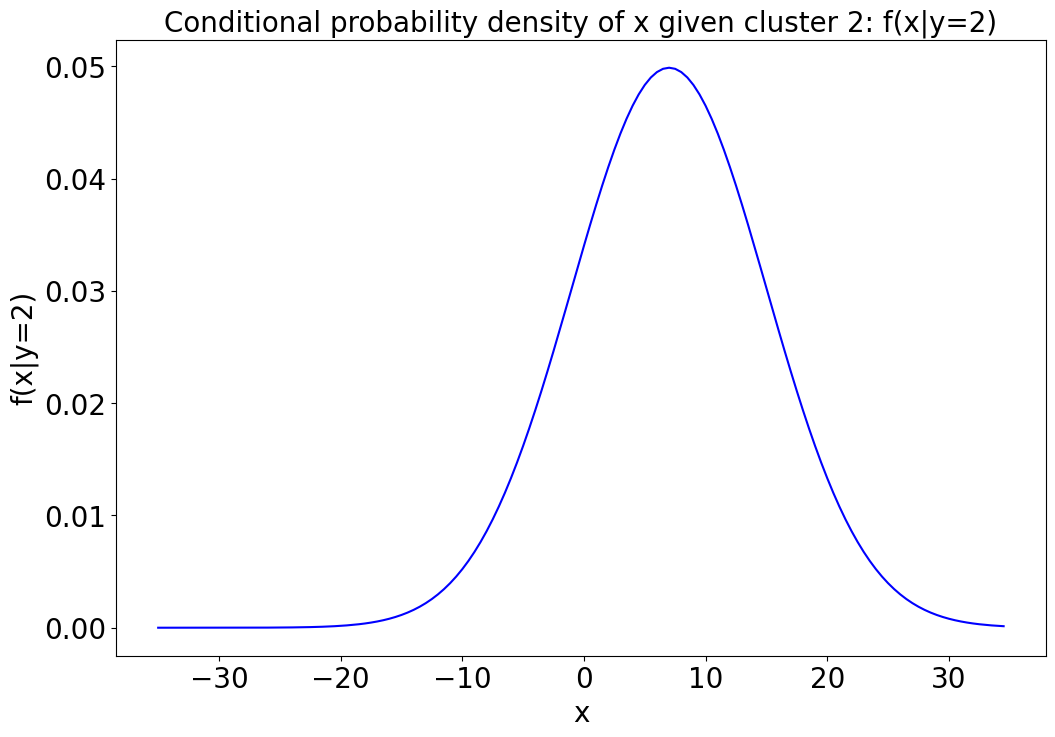}\\
\mbox{\small{(a)}} & \mbox{\small{(b)}} \\
\includegraphics[width = 0.48\linewidth]{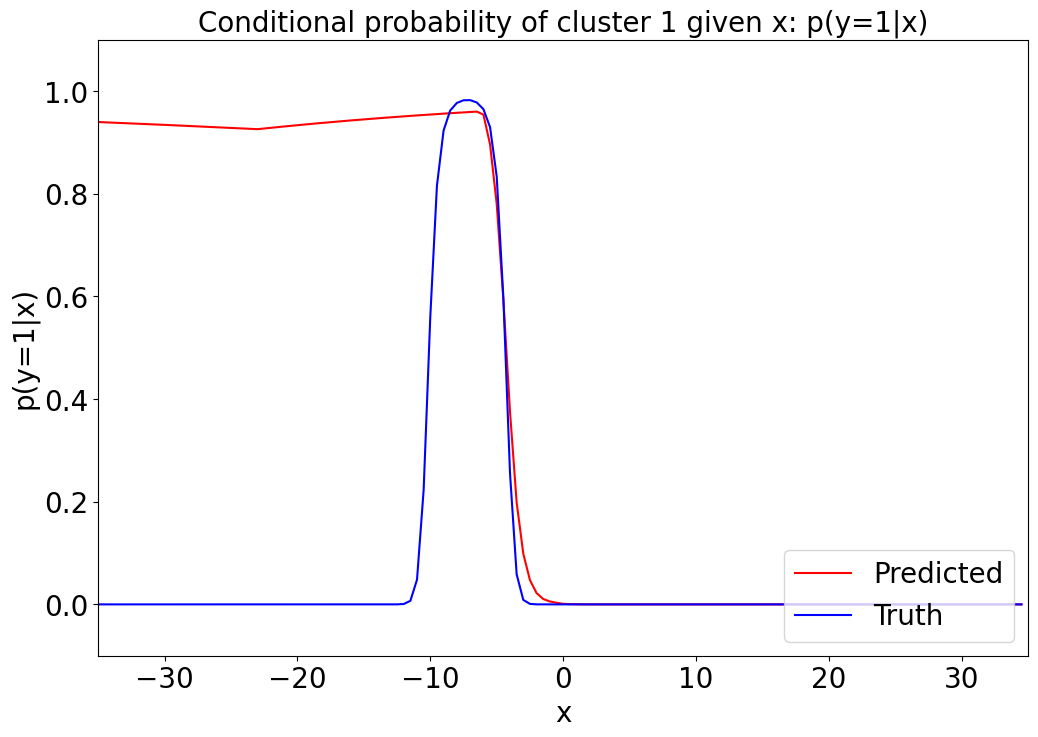} &
\includegraphics[width = 0.48\linewidth]{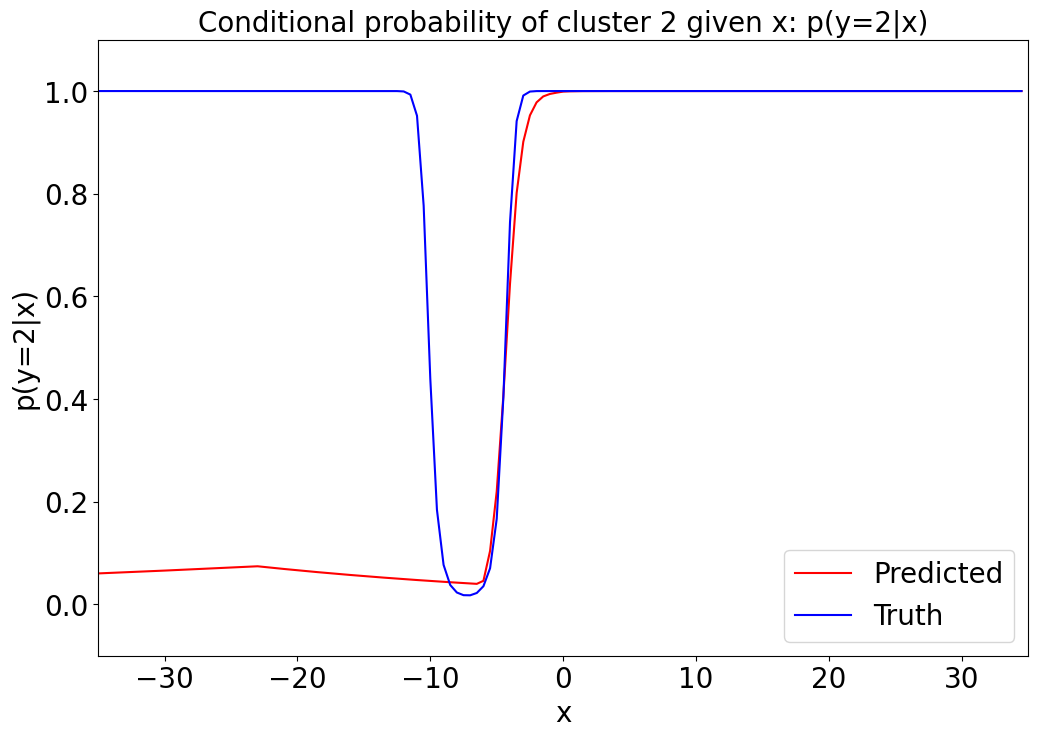}\\
\mbox{\small{(c)}} & \mbox{\small{(d)}}\\
\includegraphics[width = 0.48\linewidth]{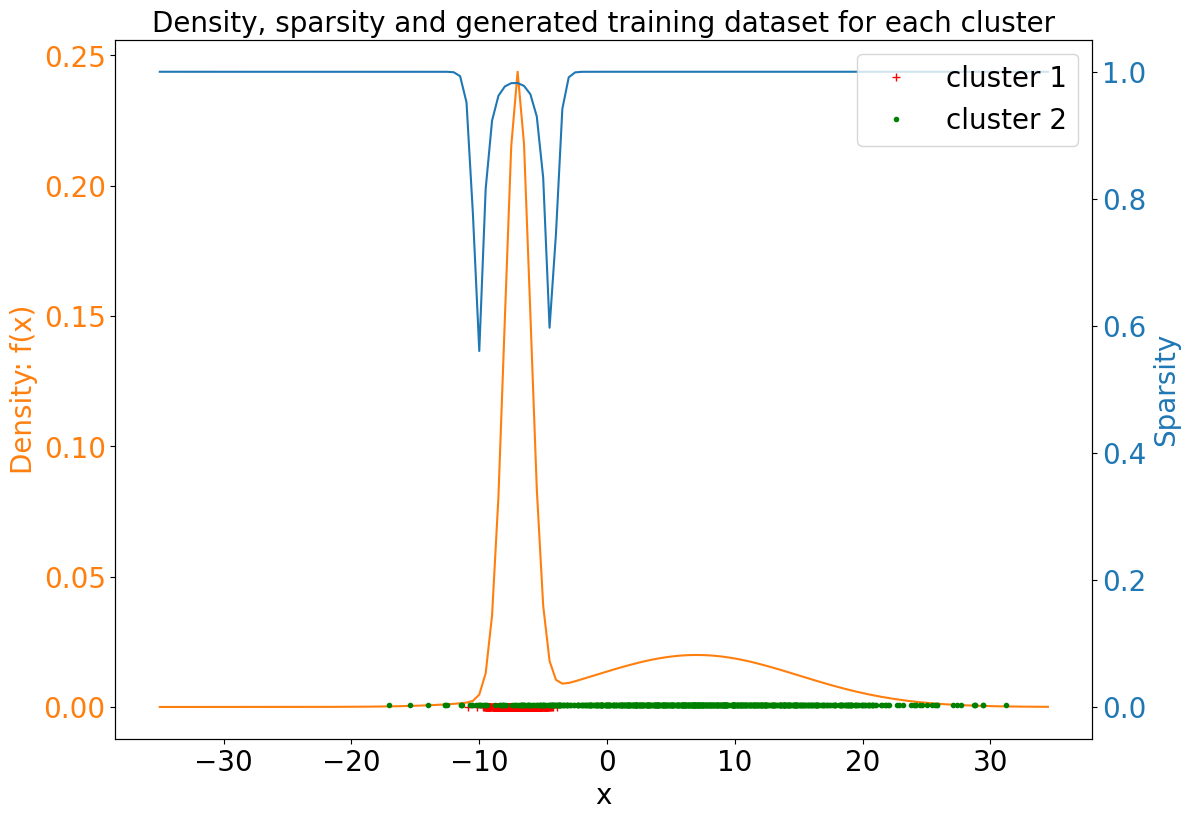} &
\includegraphics[width = 0.48\linewidth]{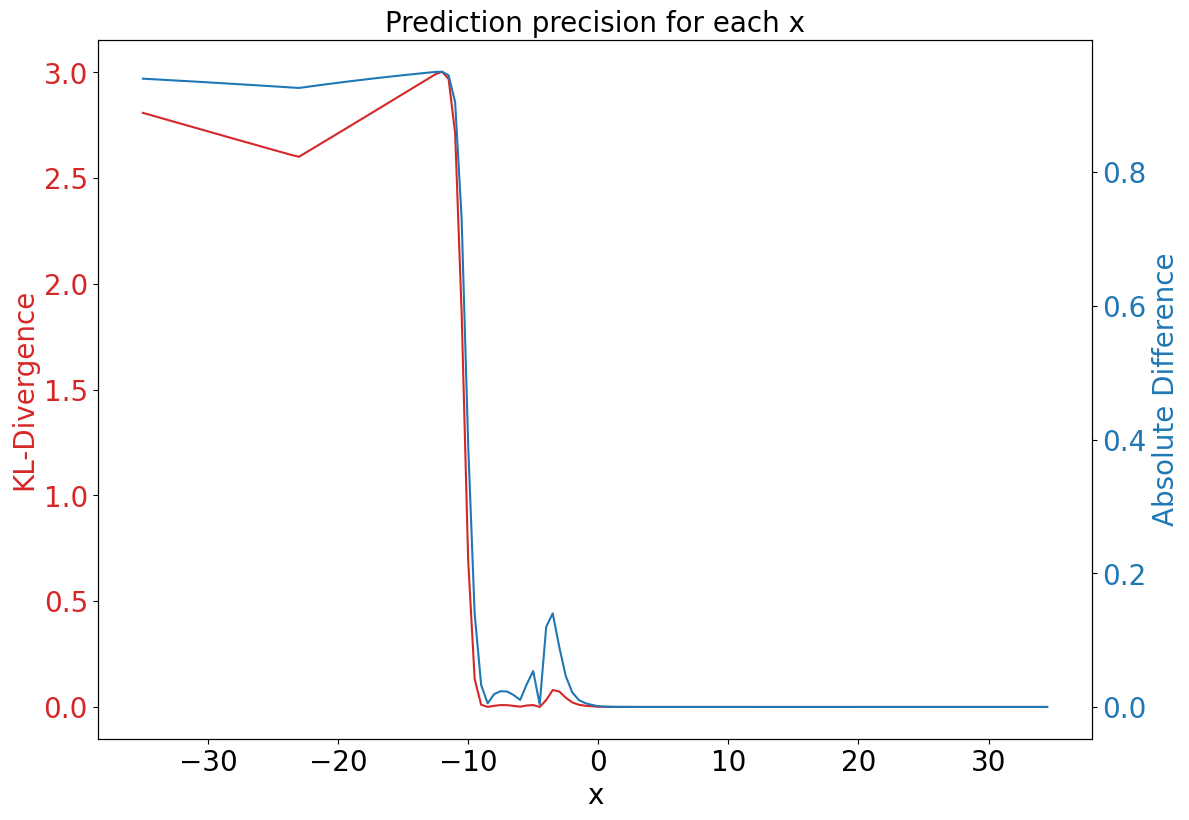}\\
\mbox{\small{(e)}} & \mbox{\small{(f)}}
\end{array}$
\end{center}
\vspace{-0.15in}
\caption{\small{Large prediction error, configurations, and results. (a) and (b) plot the probability density of clusters 1 and 2: $f(x|y=1)$ and $f(x|y=2)$, respectively. (c) and (d) plot the predicted and true $p(y|x)$ for clusters 1 and 2, respectively. We can see a large discrepancy happens when $x<-10$. (e) plots the density, sparsity, and generated training dataset for both clusters. (f) plots the prediction precision in KL-Divergence and Absolute Difference. (e) shows the density drops and the sparsity increases drastically when $x<-10$, coinciding with the sudden increase of the prediction error in (f).}}
\label{fig:1d_large_discrepancy}
\end{figure}

\begin{figure}[t!]
\begin{center}
$\begin{array}{c@{\hspace{0.03\linewidth}}c}
\includegraphics[width = 0.48\linewidth]{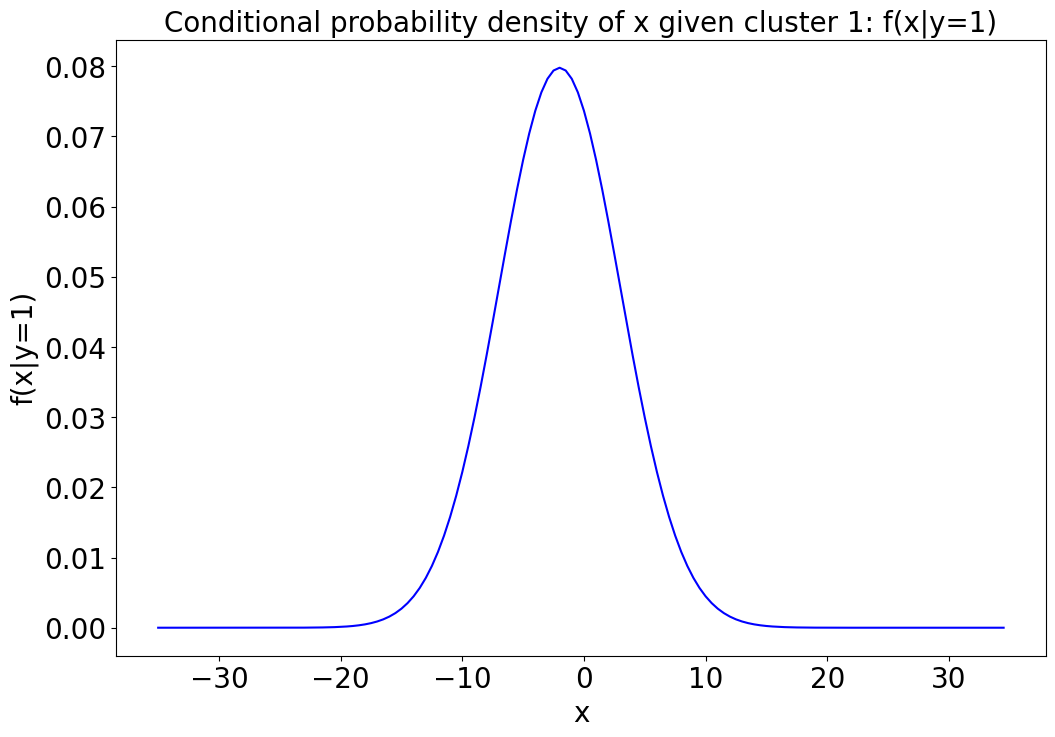} &
\includegraphics[width = 0.48\linewidth]{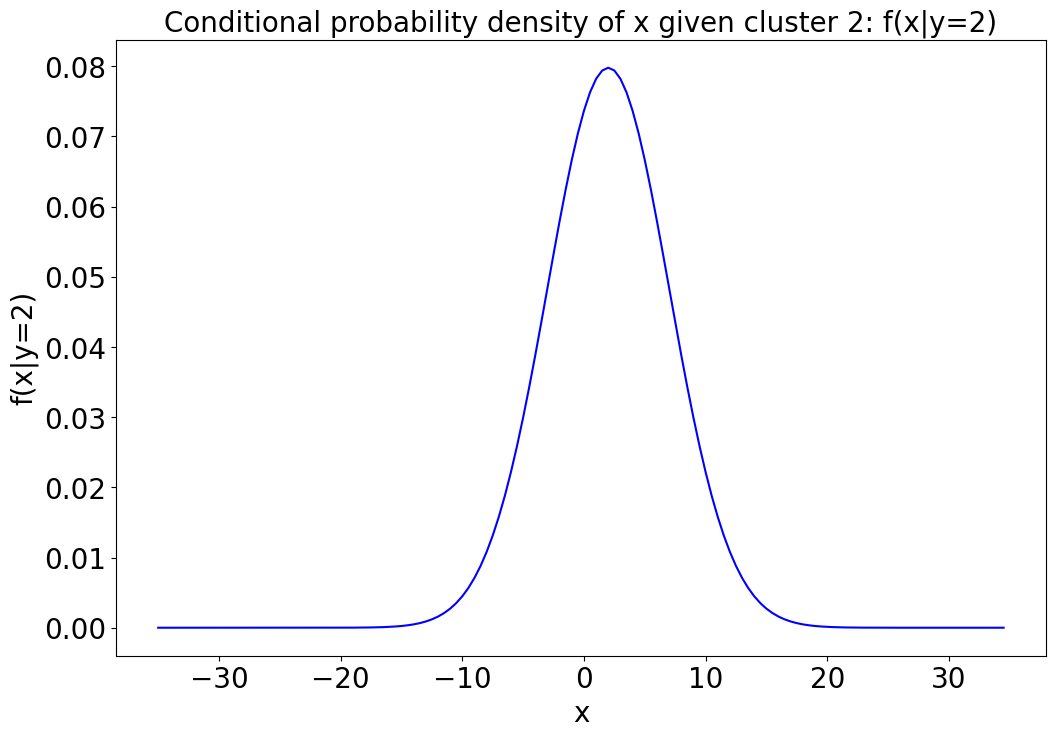}\\
\mbox{\small{(a)}} & \mbox{\small{(b)}} \\
\includegraphics[width = 0.48\linewidth]{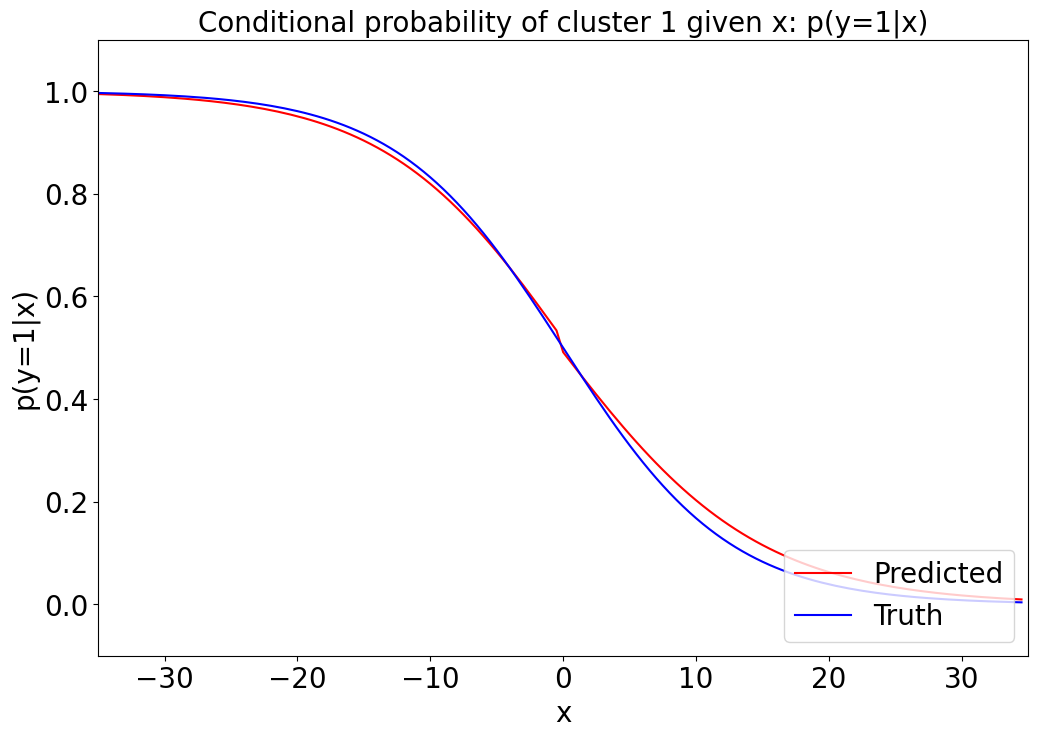} &
\includegraphics[width = 0.48\linewidth]{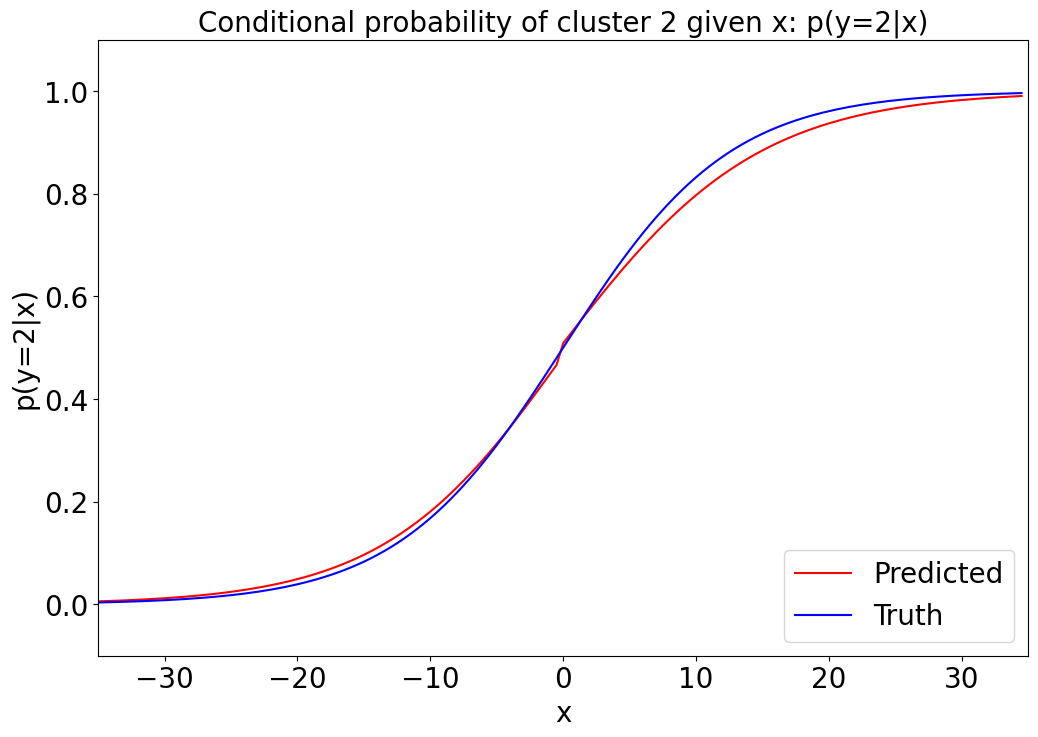}\\
\mbox{\small{(c)}} & \mbox{\small{(d)}}\\
\includegraphics[width = 0.48\linewidth]{pics/1d_small_discrepancy_p_x_y_2.png} &
\includegraphics[width = 0.48\linewidth]{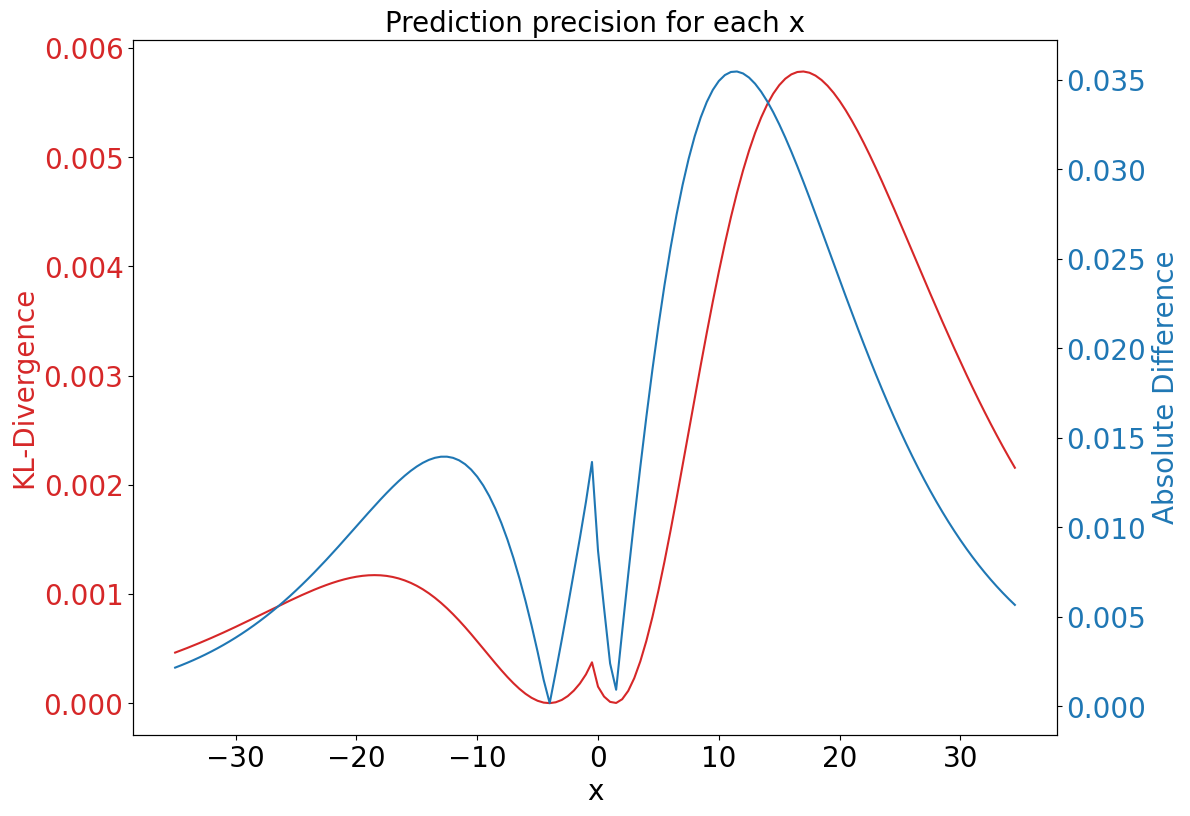}\\
\mbox{\small{(e)}} & \mbox{\small{(f)}}
\end{array}$
\end{center}
\vspace{-0.15in}
\caption{\small{Small prediction error: configurations and results. (a) and (b) plot the probability density of clusters 1 and 2: $f(x|y=1)$ and $f(x|y=2)$, respectively. (c) and (d) plot the predicted and true $p(y|x)$ for clusters 1 and 2, respectively. (e) plots the density, sparsity, and generated training dataset for both clusters. (f) plots the prediction precision in KL-Divergence and Absolute Difference. (c) shows the sparsity is high and density is low for $x<-20$ and $x>20$, but the prediction error in (f) stays low for all $x$.}}
\label{fig:1d_small_discrepancy}
\end{figure}

\paragraph{Considering Density and Sparsity as Influencing Factors}
Second, for a grid point (i.e., a parametric configuration), we record the prediction precision, together with $f(x)$ and $p(y=k)\phi_k(x)$ ($k=1,2$) sampled from $x \in [-35,35]$ with a step size of $0.5$. Thus, we collect $9000\times142=1278000$ points in total. As we define the prediction precision at each sampled $x$ using KL-Divergence and Absolute Difference, we can plot the scattered points illustrating the relationship between the prediction precision and two extra potential influencing factors: density and sparsity, which are defined as $\small{
Density(x)=f(x)}$ and $\small{Sparsity(x)=\frac{|p(y=1)\phi_1(x)-p(y=2)\phi_2(x)|}{p(y=1)\phi_1(x)+p(y=2)\phi_2(x)}}$, respectively. Sparsity measures how much each cluster relatively contributes to the whole density. Fig. \ref{fig:1d_density_sparsity} illustrates the prediction precision as the function of density and sparsity. We can see that the prediction precision roughly obeys the power-law with a large exponent for both density and sparsity. The prediction precision decreases drastically as the density increases while it increases drastically as the sparsity increases. This means that \textbf{most prediction failures are observed when the density is low AND the sparsity is high.} This condition is usually satisfied at the far outer side of both clusters or in between two clusters when their variances are low. With this observation and revisiting Fig. \ref{fig:1d_grid_search} (c) and (d), we can see that when $\mu_1$ increases, the mean $\mu_2=-\mu_1$ of cluster 2 decreases. Thus, as these two clusters' distance increases, it becomes easier to satisfy the failure condition at a certain sampled $x$. Similarly, in Fig. \ref{fig:1d_grid_search} (e) and (f), when the variances of both clusters are small, the failure condition tends to happen more frequently.

\subsubsection{Example Cases with Specific Configurations}

To further illustrate our observation in Section \ref{sec:systematic_analysis},  we select two specific configurations with large and small prediction errors from the results in Fig. \ref{fig:1d_density_sparsity}. For each configuration, we follow our assessment framework: generate a training dataset, train a DNN-based classifier, and compare the prediction made by the trained classifier with the truth induced by our data generator.

\paragraph{Configuration with Large Prediction Error}

In this configuration,  for cluster 1, we have its mean as $-7$, variance as $1$, and mixing coefficient as $0.6$. For cluster 2, we have its mean as $7$, variance as $8$, and mixing coefficient as $0.4$.
Fig. \ref{fig:1d_large_discrepancy} shows the parametric configurations and experiment results for this setting. We can see that large prediction errors occur when $x<-10$, which coincides with low density and high sparsity. This observation is in accordance with the results from our grid search, which showed most prediction failures are observed with low density and high sparsity.

\begin{figure*}[t!]
    \centering

    \begin{subfigure}[h]{0.25\textwidth}
        \includegraphics[width=\textwidth]{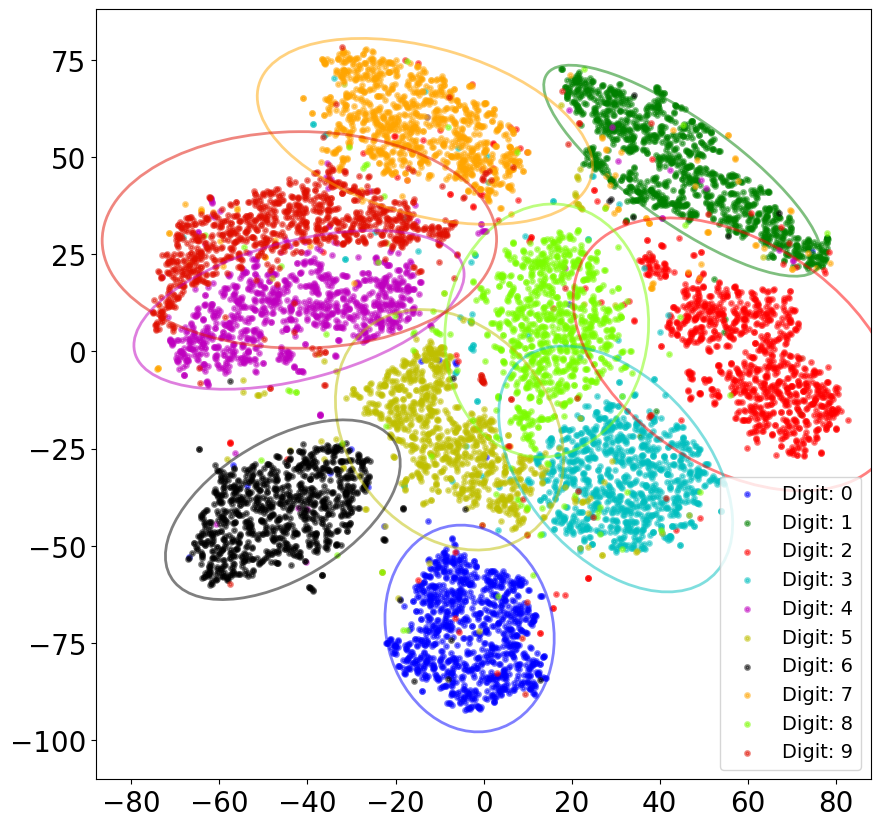}
        \caption{}
    \end{subfigure}
    \begin{subfigure}[h]{0.25\textwidth}
        \includegraphics[width=\textwidth]{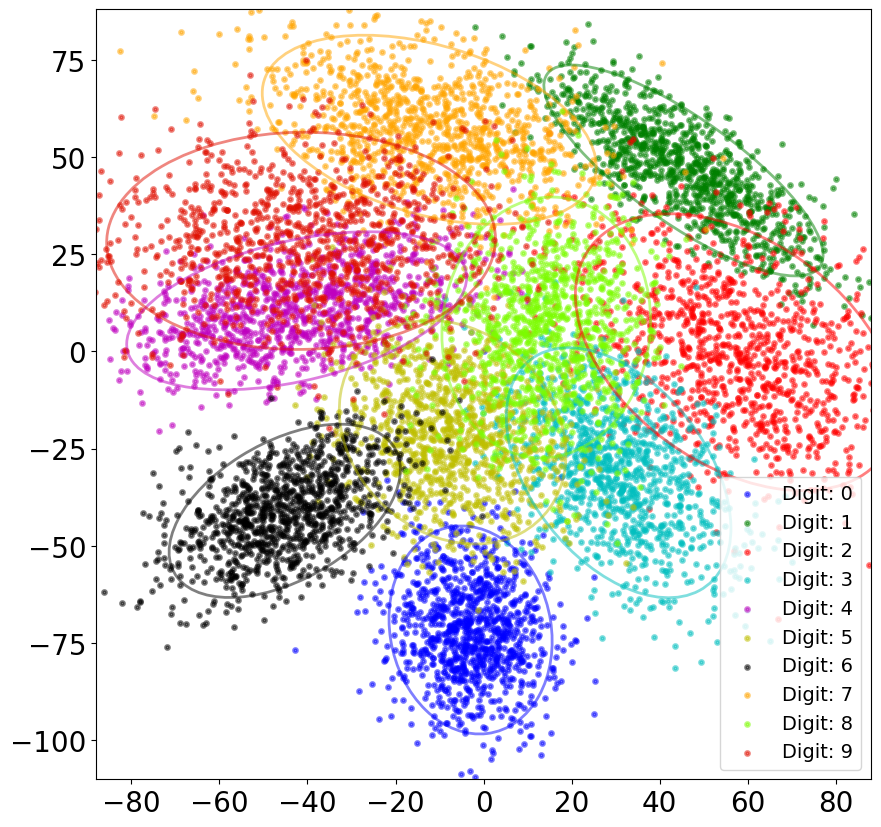}
        \caption{}
    \end{subfigure}
    \begin{subfigure}[h]{0.42\textwidth}
        \includegraphics[width=\textwidth]{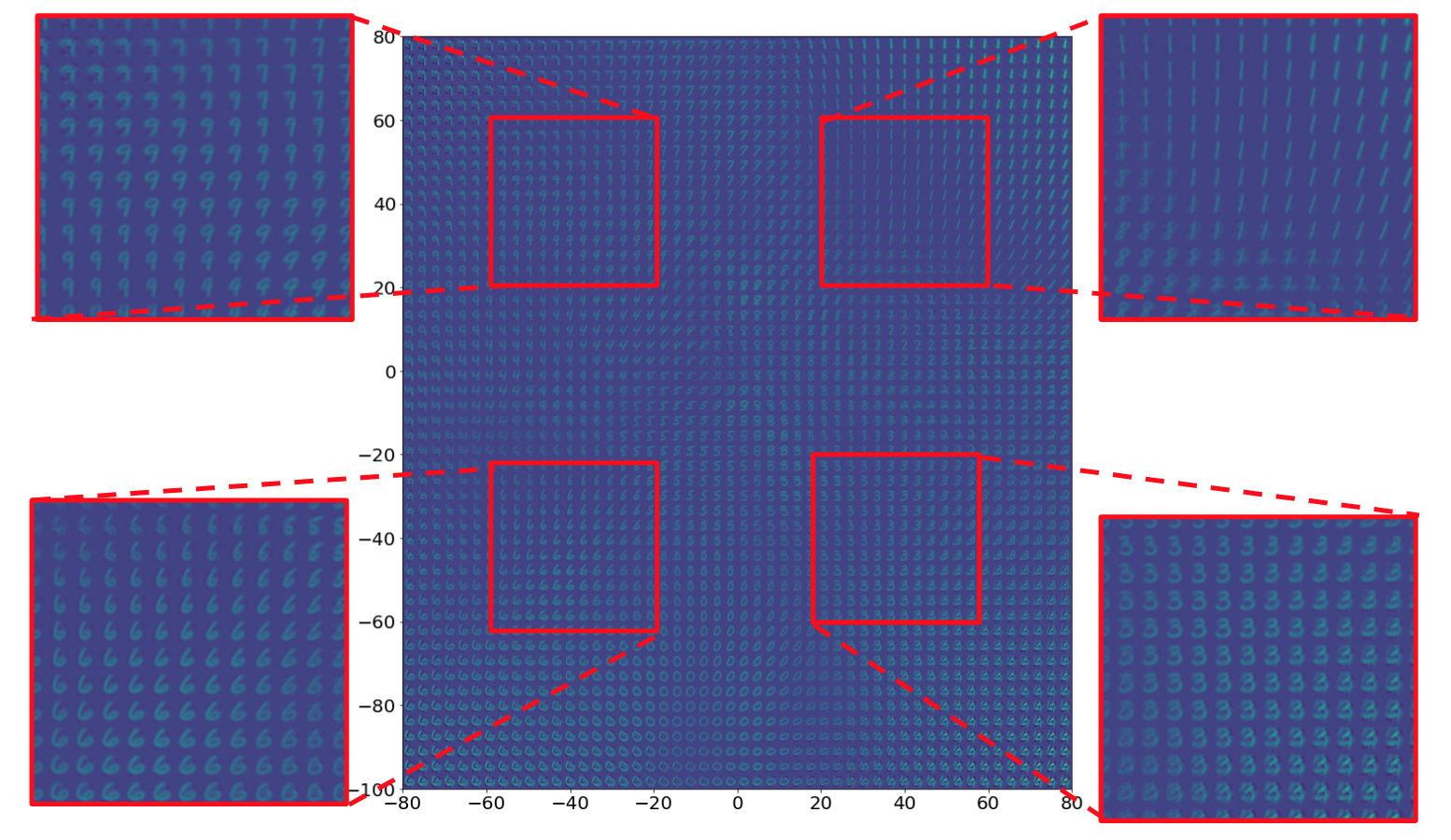}
        \caption{}
    \end{subfigure}
  \vspace{-0.1in}
  \caption{\small{2-D data and digit planes. (a) shows the t-SNE dimension reduction result of the MNIST training dataset on a 2-D plane where a elliptical circle surrounds each category and shows the area covered by $2\sigma$ of the data samples for each digit. (b) shows the synthetic training data for each category on the 2-D plane. (c) shows the corresponding digit for each grid point on the 2-D plane.}}
  \label{fig:2d_data_plane}
\end{figure*}

\begin{figure*}[t!]
    \centering
    \begin{subfigure}[h]{0.9\textwidth}
        \includegraphics[width=\textwidth]{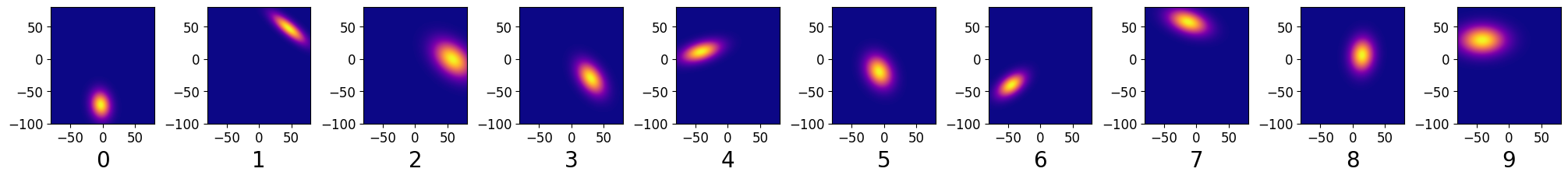}
    \end{subfigure}
    \begin{subfigure}[h]{0.9\textwidth}
         \includegraphics[width=\textwidth]{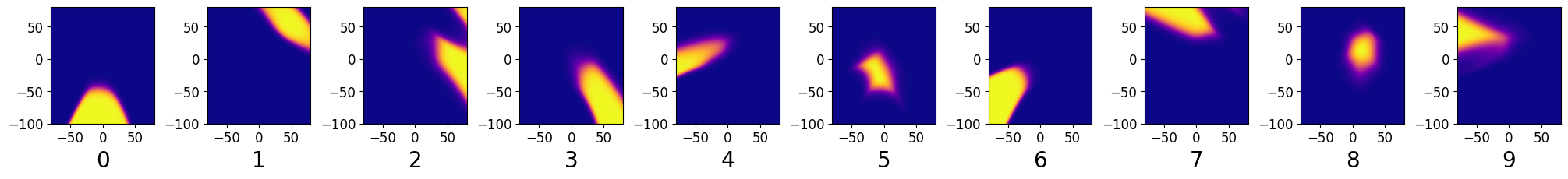}
    \end{subfigure}
    \begin{subfigure}[h]{0.9\textwidth}
         \includegraphics[width=\textwidth]{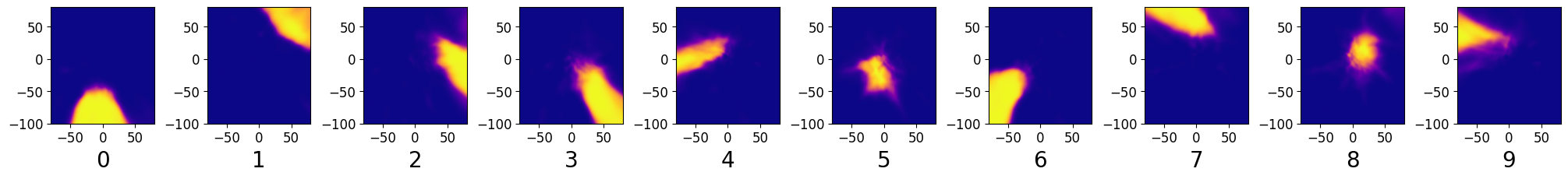}
    \end{subfigure}
    \begin{subfigure}[h]{0.9\textwidth}
         \includegraphics[width=\textwidth]{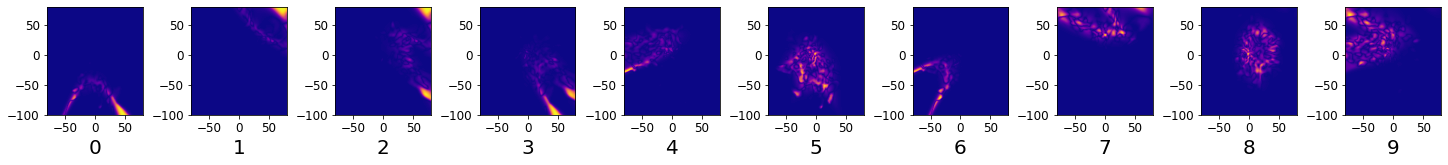}
    \end{subfigure}
    \vspace{-0.1in}
    \caption{\small{$f(\mathbf{v}|y)$, true $p(y|\mathbf{v})$, and predicted $p(y|\mathbf{v})$ shown on 2-D plane. The first row shows the distribution density $f(\mathbf{v}|y)$ for each digit category. The second and third rows illustrate the true posterior probability $p(y|\mathbf{v})$ and the predicted $p(y|\mathbf{v})$, respectively. The last row shows the pixel-wise absolute difference between the second and third rows. }}
    \label{fig:2d_p_x_y_and_p_y_x}
\end{figure*}

\paragraph{Configuration with Small Prediction Error}

In this configuration, we have the mean values of two clusters at $-2$ and $2$, respectively. Each cluster has a variance of $5$ and a mixing coefficient of 0.5. Fig. \ref{fig:1d_small_discrepancy} shows the parametric configurations and experiment results for this setting. We can see that even though the regions of $x<-20$ and $x>20$ satisfy low density and high sparsity, the prediction error stays low. This observation does not, however, contradict the experiment results from our grid search, because low density and high sparsity are necessary but not sufficient conditions for high prediction errors.

\begin{figure*}[t]
    \centering

    \begin{subfigure}[h]{0.225\textwidth}
        \includegraphics[width=\textwidth]{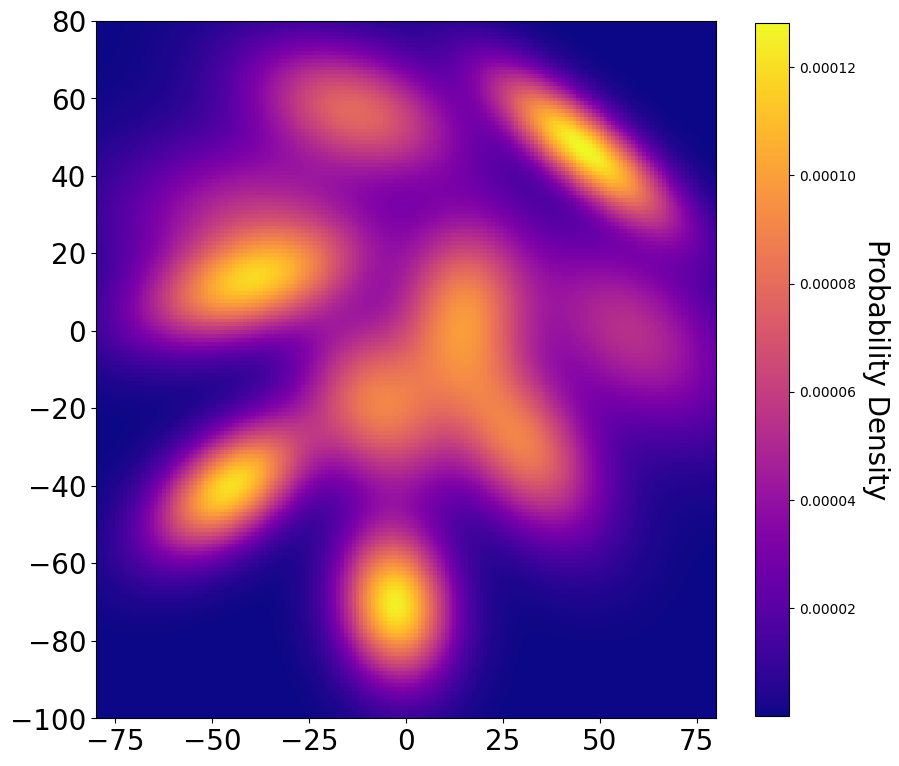}
        \caption{}
    \end{subfigure}
      \begin{subfigure}[h]{0.225\textwidth}
        \includegraphics[width=\textwidth]{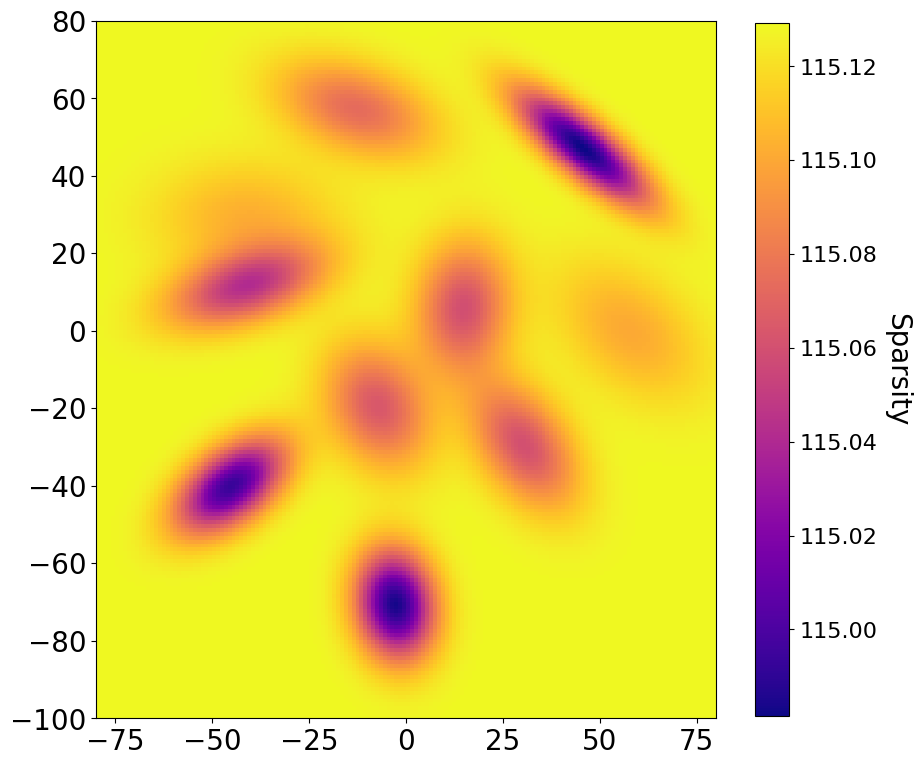}
        \caption{}
    \end{subfigure}
    \begin{subfigure}[h]{0.225\textwidth}
        \includegraphics[width=\textwidth]{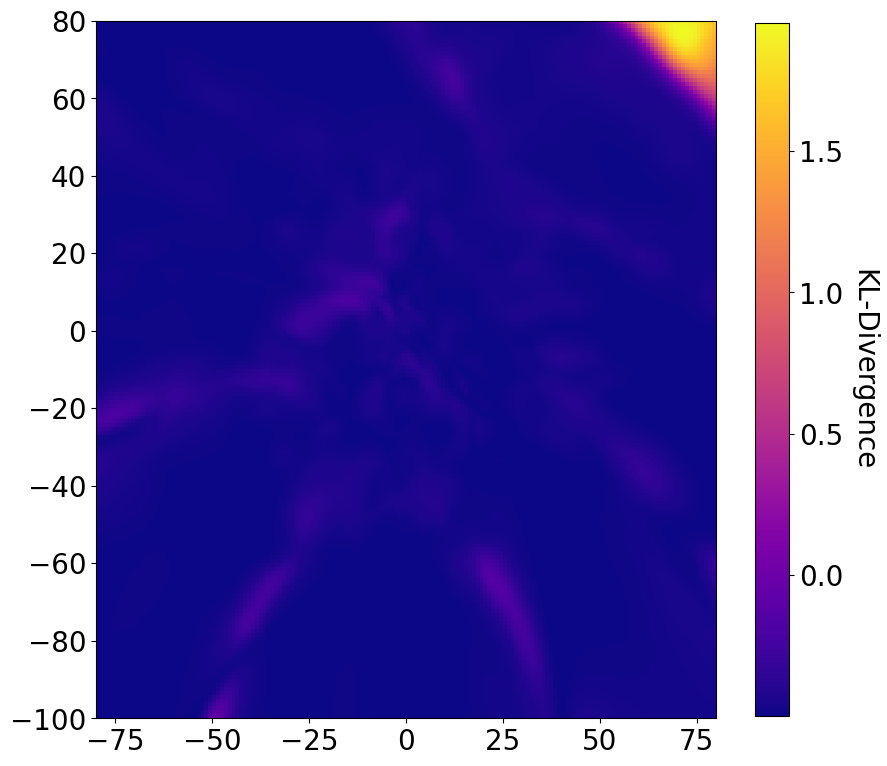}
        \caption{}
    \end{subfigure}
    \begin{subfigure}[h]{0.225\textwidth}
        \includegraphics[width=\textwidth]{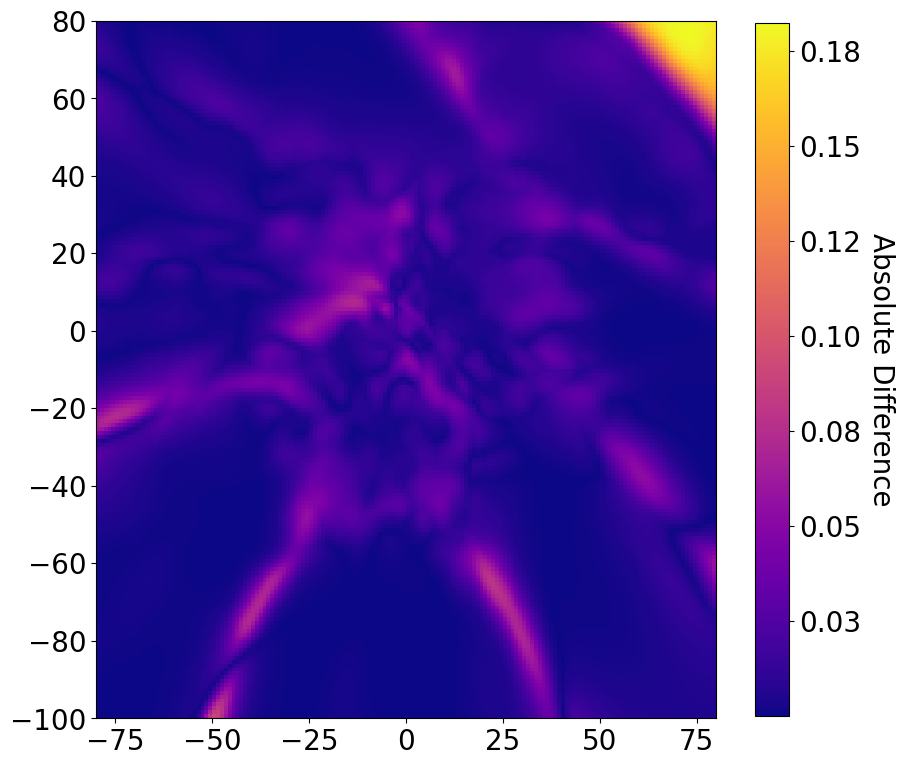}
        \caption{}
    \end{subfigure}
   \vspace{-0.1in}
  \caption{\small{Density, sparsity, and prediction precision on 2-D plane. (a) shows the probability density $f(\mathbf{v})$. (b) shows the sparsity for each 2-D data point $\mathbf{v}$. (c) and (d) show the prediction error in KL-Divergence and Absolute Difference on the 2-D plane, respectively.}}
  \label{fig:2d_density_sparsity_precision}
\end{figure*}

\begin{figure*}[t]
    \centering
    \begin{subfigure}[h]{0.225\textwidth}
        \includegraphics[width=\textwidth]{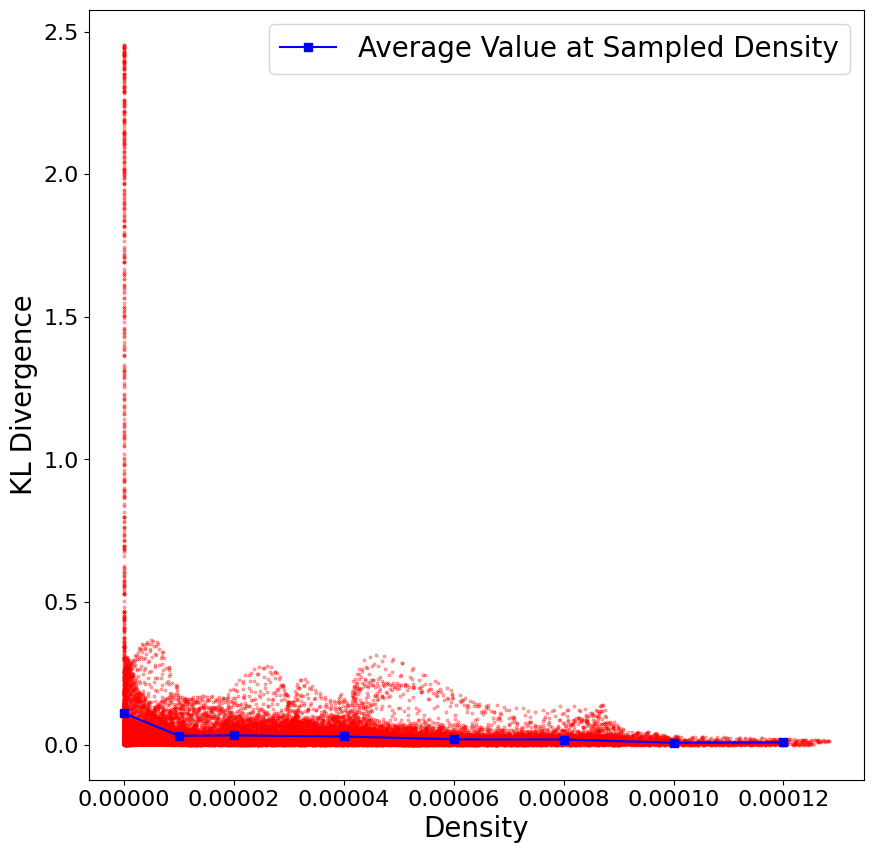}
        \caption{}
    \end{subfigure}
    \begin{subfigure}[h]{0.225\textwidth}
        \includegraphics[width=\textwidth]{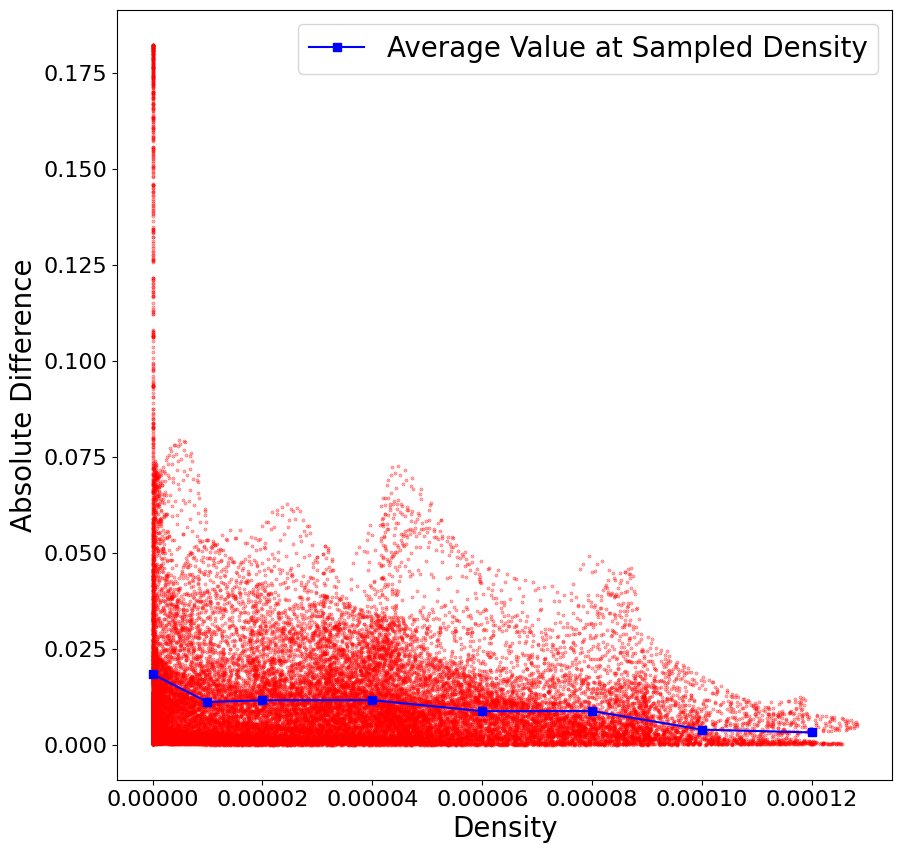}
        \caption{}
    \end{subfigure}
    \begin{subfigure}[h]{0.225\textwidth}
        \includegraphics[width=\textwidth]{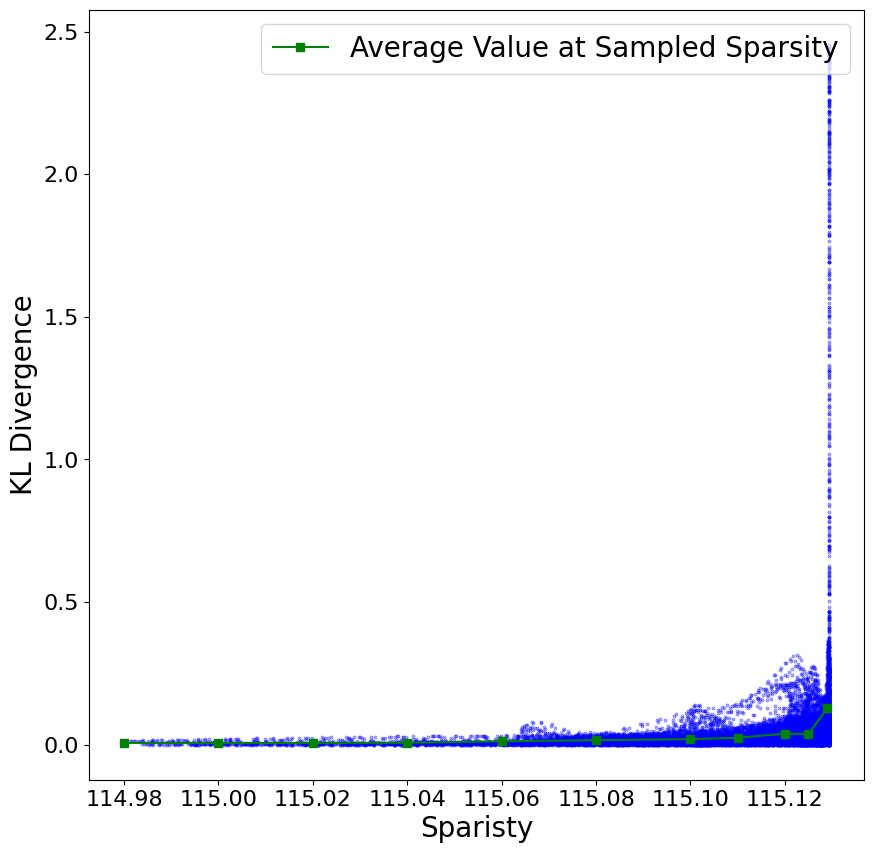}
        \caption{}
    \end{subfigure}
    \begin{subfigure}[h]{0.225\textwidth}
        \includegraphics[width=\textwidth]{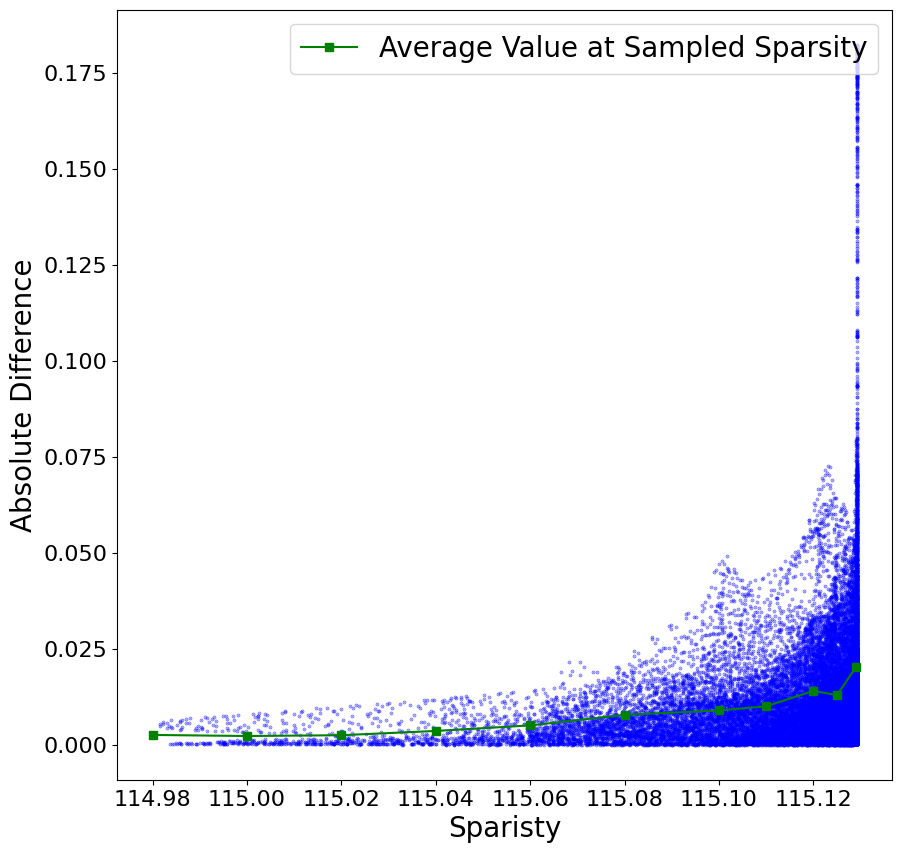}
        \caption{}
    \end{subfigure}
   \vspace{-0.1in}
  \caption{\small{Prediction precision as the function of density and sparsity. (a) and (b) show the density's impact on the prediction precision in terms of KL-Divergence and Absolute Difference. (c) and (d) show the sparsity's influence on the prediction precision in terms of KL-Divergence and Absolute Difference. For each plot, we also divide the x axis into several bins and calculate the average value for each bin. Then, we overlay the average values as the function of sampled bins on each figure.}}
  \label{fig:2d_precision_as_function_of_density_and_sparsity}
\end{figure*}

\subsection{Pseudo High Dimensional Case}
\label{sec:hd_case}

To investigate whether these conclusions continue to hold in high dimensional cases, we start as in the 1-D case but using mixture of ten 2-D Gaussians as the data generator. After the random samples on the 2-D plane are acquired, we use a reconstructing mapping function $f_{re}:\mathbf{V} \subset \mathbb{R}^2 \rightarrow \mathbf{X} \subset \mathbb{R}^d$ to map a 2-D random sample $\mathbf{v}$ to a $d$-dimensional sample $\mathbf{x}$. The high dimensional generative model can be parameterized similar to the 1-D case in Section \ref{sec:1d_experiment}: $f(\mathbf{v})=\sum_{k=0}^{9} p(y=k)\phi_k(\mathbf{v})$, where $\sum_{k=0}^{9} p(y=k)=1$, $\phi_k(\mathbf{v})=\mathcal{N}(\boldsymbol{\mu}_k,\boldsymbol{\Sigma}_k)$, and $\mathbf{x}=f_{re}(\mathbf{v})$.

We get $f_{re}$ by reversing the t-SNE~\cite{van2008visualizing} dimension reduction of the MNIST dataset\cite{lecun-mnisthandwrittendigit-2010}. We first apply t-SNE on the MNIST dataset to obtain a set of data samples on a 2-D manifold and train a DNN to map these 2-D samples back to the original MNIST images. Thus, this DNN acts as our $f_{re}$. We illustrate the training process of $f_{re}$ in more detail in Appendix \ref{sec:proof}.
Since we consider the MNIST dataset embedded in a 2-D manifold and the DNN is continuous, we assume our $f_{re}$ is smooth and bijective. Here, we could also use other generative functionsm, such as the decoder part of an autoencoder or GAN, as $f_{re}$. Still, as shown in Fig. \ref{fig:2d_data_plane} (a), a single Gaussian pdf for each digit category on the 2-D plane acquired by the t-SNE reduction is adequate. The first row of Fig. \ref{fig:2d_p_x_y_and_p_y_x} illustrates the conditional distribution $f(\mathbf{v}|y)$ for each categorical variable $y$. In our experiment, we set each $p(y=k)=0.1$. Thus, we can see that the marginal density function $f(v)$ in Fig. \ref{fig:2d_density_sparsity_precision} (a) is a simple addition of the first row of Fig. \ref{fig:2d_p_x_y_and_p_y_x} (a). Once we calculate all the parameters for each Gaussian pdf, we can generate our training dataset as shown in Fig. \ref{fig:2d_data_plane} (b), and use $f_{re}$ to map the 2-D samples back to the original image space. Fig. \ref{fig:2d_data_plane} (c) shows how each 2-D point maps to the original image space.

After generating the synthetic training dataset with $60000$ samples, we train a convolutional neural network to classify the digits. The second and third rows of Fig. \ref{fig:2d_p_x_y_and_p_y_x} show the true $p(y|\mathbf{v})$ and the predicted $p(y|\mathbf{v})$ on the 2-D plane, respectively. From the last row of Fig. \ref{fig:2d_p_x_y_and_p_y_x}, we can see that the prediction is generally accurate from the shape of the light colored area for each digit.

Similarly to Section \ref{sec:1d_experiment}, we want to find the influencing factors of prediction precision. Here, we focus on density and sparsity. As it is difficult to estimate the actual $f(\mathbf{x})$ and $f(\mathbf{x}|y)$ when considering density and sparsity, we use $f(\mathbf{v})$ and $f(\mathbf{v}|y)$ instead. Thus, the density is $f(\mathbf{v})$ and we adopt $H_G$ sparsity~\cite{hurley2009comparing}:
\begin{equation}
\label{eq:HG_sparsity}
\small{
H_G(\mathbf{v})=-\sum_k\log(p(y=k)*\phi_k(\mathbf{v}))
}
\end{equation}
We still use KL-Divergence and Absolute Difference as the metrics of prediction precision. Fig. \ref{fig:2d_density_sparsity_precision} shows density, sparsity, and prediction precision on the 2-D plane, and their correlation. Generally speaking, the areas of high prediction error (i.e., the light-colored area in (c) and (b)) correspond to the areas of low density and high sparsity. Fig. \ref{fig:2d_precision_as_function_of_density_and_sparsity} further shows the prediction precision as the function of density and sparsity. Again, we get a similar conclusion as 1-D cases that the prediction precision follows the power law as the function of density and sparsity. We choose three 1-D paths on the 2-D plane to illustrate our conclusion in detail in Appendix \ref{sec:proof}.

\section{Conclusion}

We design an innovative framework for characterizing the uncertainty associated with a DNN's approximation to the conditional distribution of its underlying training dataset. We develop a two-path evaluation paradigm in that we use Bayesian inference to obtain the theoretical ground truth based on a generative model and use the sampling and training to acquire a DNN based classifier. We compare the prediction made by the fully trained DNN with the theoretical ground truth and evaluate its capability as a probability distribution estimator. We conduct extensive experiments for 1-D and high-dimensional cases. For both cases, we come to similar conclusions that \textbf{most prediction failures made by DNN are observed when the (local) data density is low and the inter-categorical sparsity is high, and the prior probability has less impacts to DNNs' classification uncertainty}. This insight may help delineate the capability of DNNs as probability estimators and aid the interpretation of the inference produced by various deep models. Interesting areas for future research include the application of our framework to more complex categorization scenarios requiring more sophisticated generative models and remappings.




\section{Acknowledgment}

This research has been sponsored in part by the National Science Foundation grant IIS-1652846.

\appendices

\section{}
\label{sec:proof}

\textbf{Proof for Section \ref{sec:bayesian_inference}:}
As we need to map a 2-D vector $\mathbf{v}$ to $d$-dimensional vector $\mathbf{x}$ using a reconstructing mapping, we assume $ f_{re}$ is a composite mapping function: $ f_{re} = f_{bi} \circ f_{iso}: \mathbf{V} \subset \mathbb{R}^2 \rightarrow \mathbf{X} \subset \mathbb{R}^d$, which is composed of a continuous bijective function $f_{bi}: \mathbf{V} \subset \mathbb{R}^2 \rightarrow \mathbf{V'} \subset \mathbb{R}^2$ and a locally isometric function $f_{iso}:\mathbf{V'} \subset \mathbb{R}^2 \rightarrow \mathbf{X} \subset \mathbb{R}^d$.
First, we can infer $p(y|\mathbf{v})$ as
\begin{equation}
\label{eq:p_y_x_2d}
\small{
p(y|\mathbf{v}) = \frac{f(\mathbf{v}|y)p(y)}{f(\mathbf{v})}=\frac{f(\mathbf{v}|y)p(y)}{\sum_y{f(\mathbf{v}|y)p(y)}}
}
\end{equation}
Then, for the continuous bijection $f_{bi}: \mathbf{V} \rightarrow \mathbf{V}'$, similarly to Eq.  \ref{eq:p_y_x_2d} we have
\begin{equation}
\label{eq:p_y_x_2d_prime}
\small{
p(y|\mathbf{v}') = \frac{f(\mathbf{v}'|y)p(y)}{f(\mathbf{v}')}
}
\end{equation}
where $v' \in \mathbb{R}^2$. As $f_{bi}$ is continuous and bijective, we have
$\small{
f(\mathbf{v}'|y) = f(\mathbf{v}|y)*|J|
}$
and
$\small{
f(\mathbf{v}') =  f(\mathbf{v})*|J|
}$, 
where $J$ is a Jaccobian determinant of $f_{bi}$. Therefore, according to Eq. \ref{eq:p_y_x_2d} and \ref{eq:p_y_x_2d_prime} we have:
\begin{equation}
\label{eq:first_eq}
\small{
p(y|\mathbf{v}) = p(y|\mathbf{v'})
}
\end{equation}
Again, for the locally isometric mapping  $f_{iso}: \mathbf{V}' \rightarrow \mathbf{X}$, for a $d$-dimensional vector $\mathbf{x}$, we have:
\begin{equation}
\label{eq:p_y_x_Dd}
\small{
p(y|\mathbf{x}) = \frac{f(\mathbf{x}|y)p(y)}{f(\mathbf{x})}
}
\end{equation}
As $f_{iso}$ is locally isometric, all probability density functions on $\mathbf{x}$ and $\mathbf{v'}$ should change proportionally. Hence, we have
$\small{
f(\mathbf{v}'|y) = f(\mathbf{x}|y)*(dx)^{d-2}
}$
and
$\small{
f(\mathbf{v}') =f(\mathbf{x})*(dx)^{d-2}
}$,
where $dx$ is one dimensional differential and $d$ is the dimension of $\mathbf{x}$.
Again, according to Eq. \ref{eq:p_y_x_2d_prime} and \ref{eq:p_y_x_Dd} we have
\begin{equation}
\label{eq:second_eq}
\small{
p(y|\mathbf{x}) = p(y|\mathbf{v}')
}
\end{equation}
Thus, we have
$\small{
p(y|\mathbf{x}) = p(y|\mathbf{v}) = \frac{f(\mathbf{v}|y)p(y)}{f(\mathbf{v})}
}$ 
based on Eq. \ref{eq:first_eq}, \ref{eq:second_eq}, and \ref{eq:p_y_x_2d}, meaning we can make use of $f(\mathbf{v}|y)$ and $p(y)$ to infer $p(y|\mathbf{x})$.


\begin{wrapfigure}{r}{0.53\linewidth}
\centering
 \includegraphics[width=1.0\linewidth]{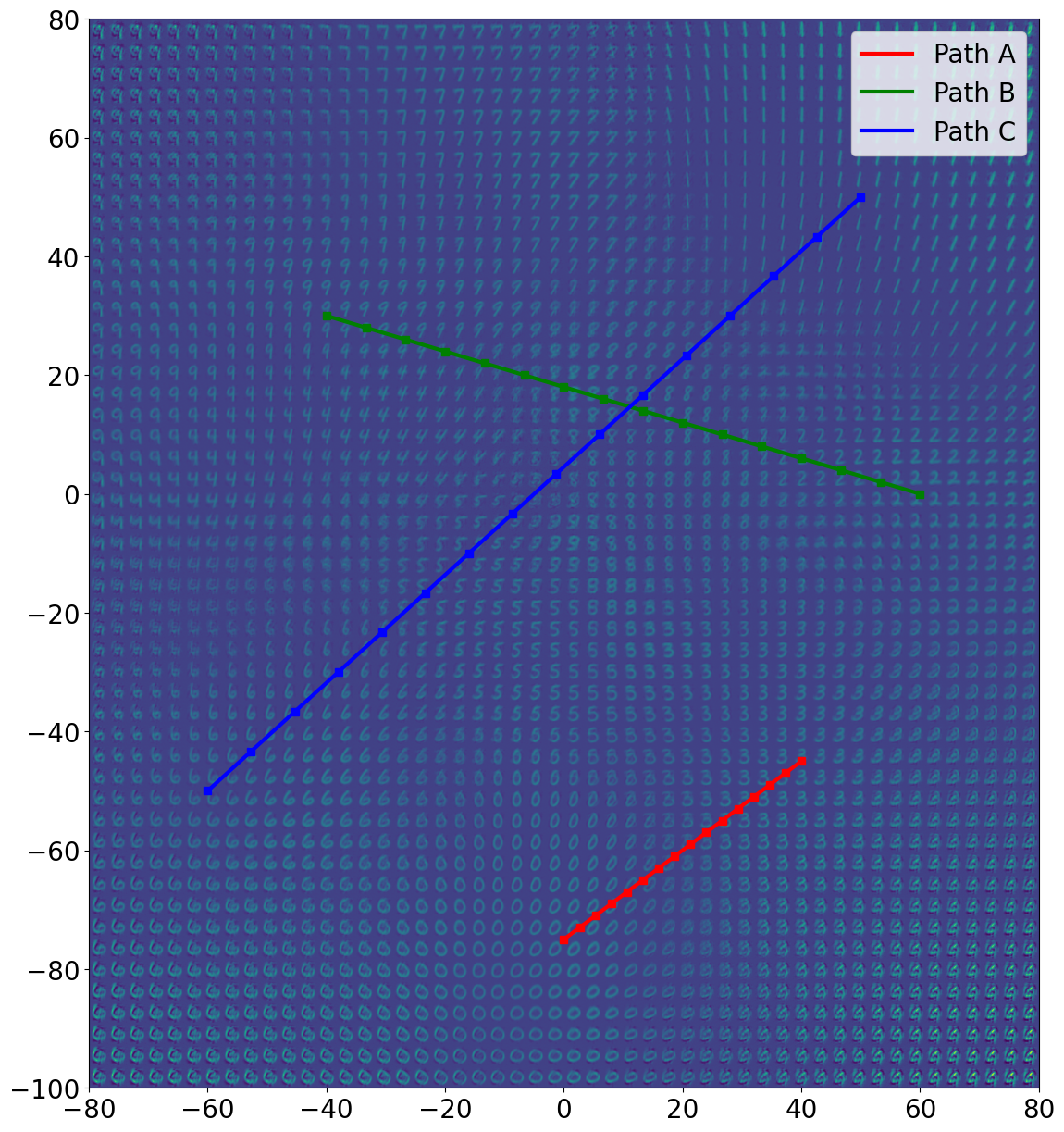}
 \vspace{-0.15in}
    \caption{\small{Three example paths.}}
 \vspace{-0.15in}
    \label{fig:paths}
\end{wrapfigure}

\begin{figure}[t]
    \centering
    \begin{subfigure}[h]{0.45\textwidth}
        \includegraphics[width=\textwidth]{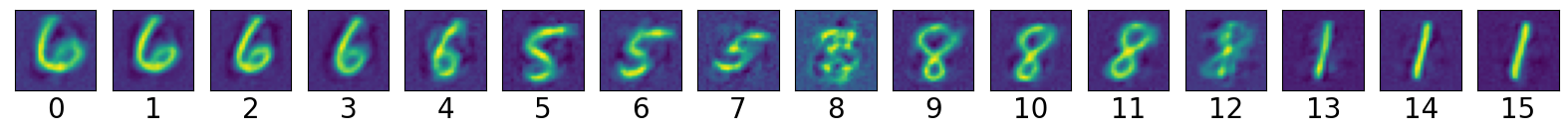}
    \end{subfigure}
    \begin{subfigure}[h]{0.45\textwidth}
         \includegraphics[width=\textwidth]{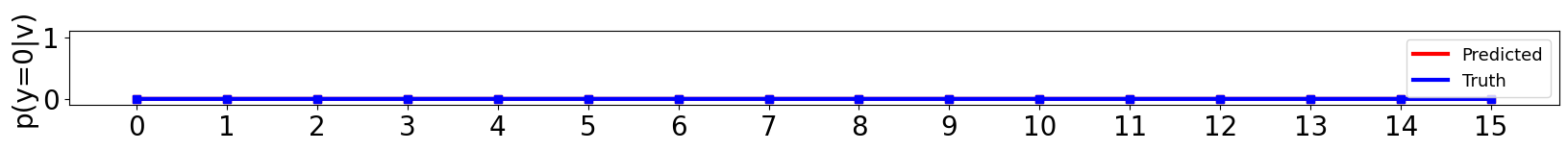}
    \end{subfigure}
    \begin{subfigure}[h]{0.45\textwidth}
         \includegraphics[width=\textwidth]{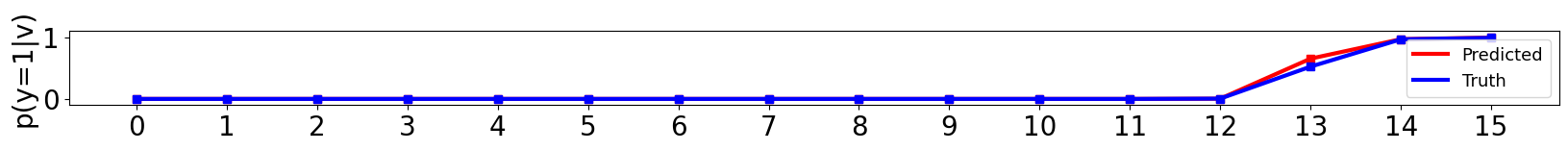}
    \end{subfigure}
    \begin{subfigure}[h]{0.45\textwidth}
         \includegraphics[width=\textwidth]{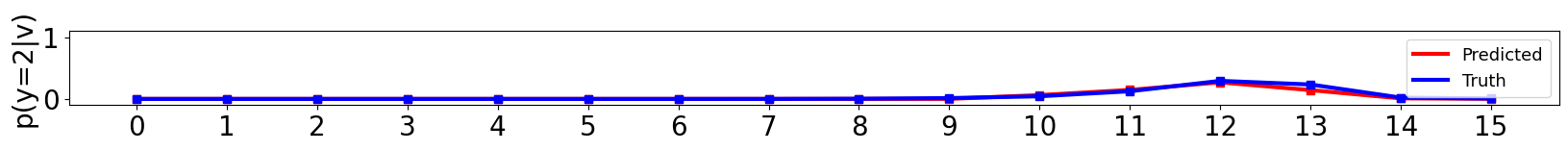}
    \end{subfigure}
    \begin{subfigure}[h]{0.45\textwidth}
         \includegraphics[width=\textwidth]{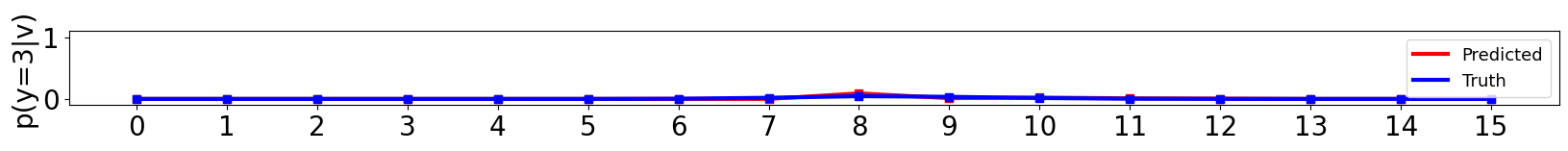}
    \end{subfigure}
    \begin{subfigure}[h]{0.45\textwidth}
         \includegraphics[width=\textwidth]{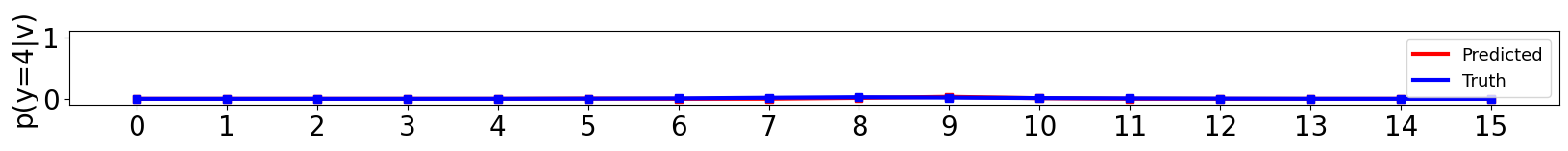}
    \end{subfigure}
    \begin{subfigure}[h]{0.45\textwidth}
         \includegraphics[width=\textwidth]{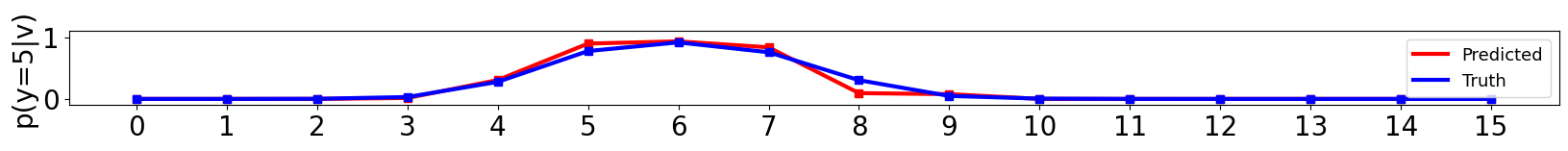}
    \end{subfigure}
    \begin{subfigure}[h]{0.45\textwidth}
         \includegraphics[width=\textwidth]{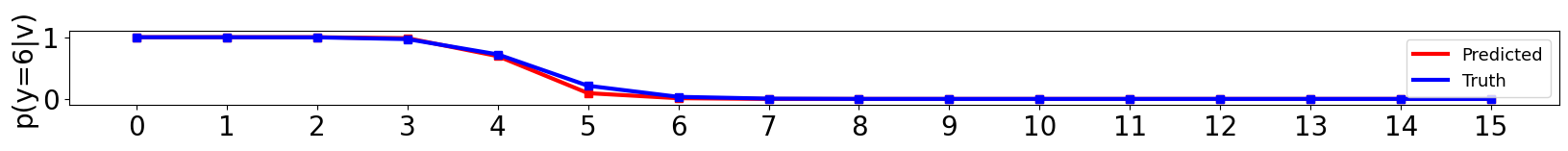}
    \end{subfigure}
    \begin{subfigure}[h]{0.45\textwidth}
         \includegraphics[width=\textwidth]{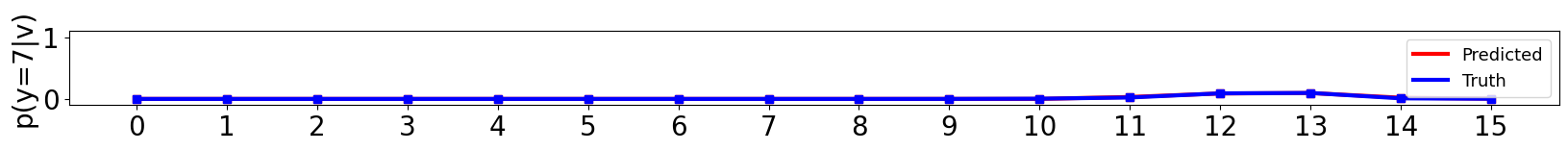}
    \end{subfigure}
    \begin{subfigure}[h]{0.45\textwidth}
         \includegraphics[width=\textwidth]{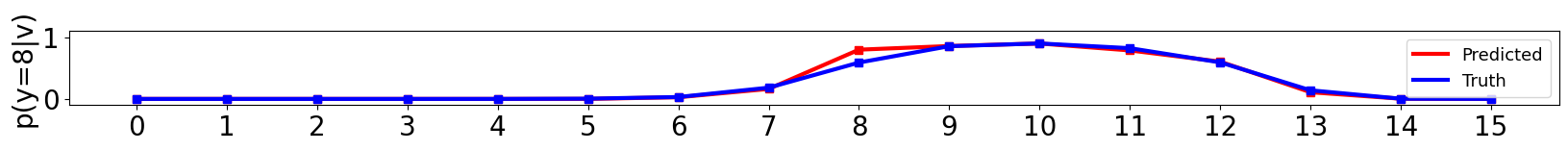}
    \end{subfigure}
    \begin{subfigure}[h]{0.45\textwidth}
         \includegraphics[width=\textwidth]{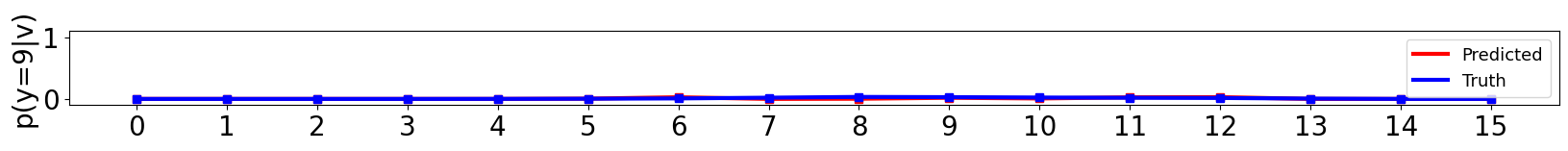}
    \end{subfigure}
    \begin{subfigure}[h]{0.45\textwidth}
         \includegraphics[width=\textwidth]{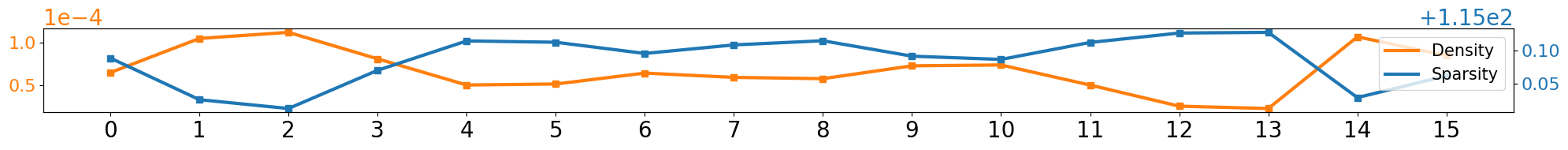}
    \end{subfigure}
    \begin{subfigure}[h]{0.45\textwidth}
         \includegraphics[width=\textwidth]{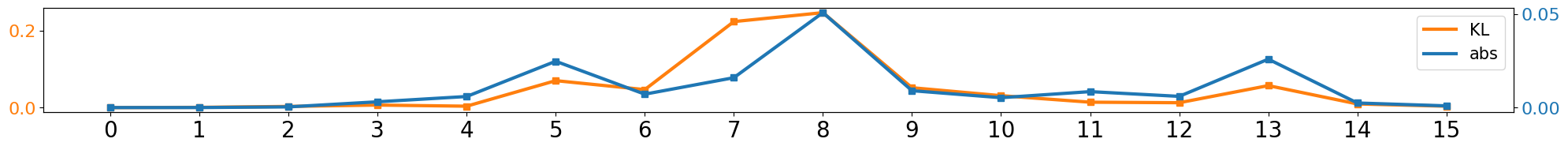}
    \end{subfigure}
    \vspace{-0.1in}
    \caption{\small{Path C goes from $6$ to $1$ through $5$ and $8$ on the digit plane with $16$ sampled images. The prediction and the theoretical truth for $p(y|\mathbf{v})$ are plotted digit-wisely. The probability $p(y=6|\mathbf{v})$ gradually decreases from 1 to 0 and $p(y=1|\mathbf{v})$ has an opposite trend. From Image $4$ to $8$, the probability of $p(y=5|\mathbf{v})$ rises and falls. From Image $7$ to $13$, the probability of $p(y=8|\mathbf{v})$ rises and falls. From Image $11$ to $13$, the probability of $p(y=2|\mathbf{v})$ fluctuates slightly, indicating these images somewhat resemble $2$. The second last row plots the density and the sparsity along this path and the last row plots the prediction precision in KL-Divergence and Absolute Difference. Three error peaks at Images $5$, $8$, and $13$, coinciding with three valleys of the density and the peaks of sparsity.}}
    \label{fig:path_3}
  \end{figure}

\textbf{1-D Paths for High Dimensional Case:}
To illustrate the conclusion about how density and sparsity correlate with prediction precision in more detail, we can select any path on the 2-D plane for high dimensional cases and show the digit-wise prediction, density, sparsity, and prediction precision along each path. Fig. \ref{fig:paths} shows three example paths on the 2-D digit plane. Fig.~\ref{fig:path_3} shows the statistics for Path C. From the second last row in Fig.~\ref{fig:path_3}, we can see that the density usually correlates with sparsity, while we assume density and sparsity should be independent of each other. We observe this phenomenon also along other paths and attribute it to the specific configuration of our 2-D Gaussian Mixture (In Section \ref{sec:comparison}, we explain the reason why we only consider a specific configuration for the high dimensional case). For 1-D, when we conduct systematical grid search in the parametric space, we remove this unwanted correlation between density and sparsity (see Fig. \ref{fig:1d_large_discrepancy} (e), in which there is no correlation between density and sparsity).

\begin{figure}[t!]
\begin{center}
\includegraphics[width = 0.9\linewidth]{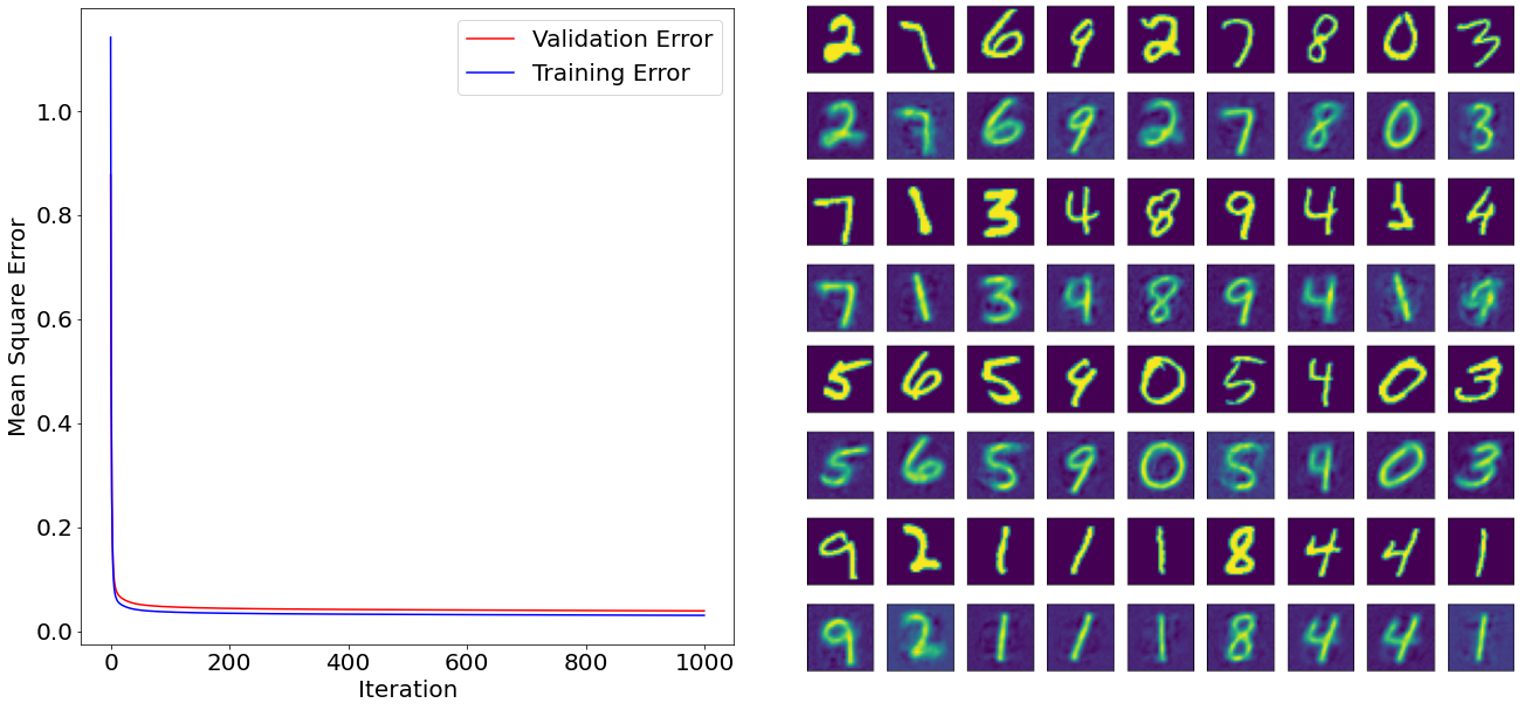}\\
\small{(a)}\hspace{1.4in}\small{(b)} 
\end{center}
\vspace{-0.15in}
\caption{\small{(a) Mean square error for training and validation of $f_{re}$ for 1000 iterations. (b) Comparison of the original MNIST digits and reconstructed digits by $f_{re}$, where the examples of original MNIST digits and the reconstructed ones are illustrated in alternating rows starting with the original.}}
\vspace{-0.1in}
\label{fig:f_re_training}
\end{figure}


\textbf{Training of $f_{re}$:}
We first apply t-SNE on the MNIST dataset to get a set of 2-D data samples and then train a fully connected DNN to map these 2-D samples back to the original MNIST images. We show the training of $f_{re}$ in Fig. \ref{fig:f_re_training} (a) and the efficacy of the reconstruction mapping in Fig. \ref{fig:f_re_training} (b).

{\small
\bibliographystyle{abbrv}
\bibliography{references}
}

\end{document}